\pdfoutput=1

\documentclass[11pt]{article}

\usepackage[final]{acl}

\usepackage{times}
\usepackage{latexsym}

\usepackage[T1]{fontenc}

\usepackage[utf8]{inputenc}

\usepackage{microtype}

\usepackage{inconsolata}

\usepackage{graphicx}
\usepackage[utf8]{inputenc} 
\usepackage[T1]{fontenc}    
\usepackage{url}            
\usepackage{booktabs}       
\usepackage{amsfonts}       
\usepackage{nicefrac}       
\usepackage{microtype}      
\usepackage{xcolor}         
\usepackage{marvosym}
\usepackage{multirow}
\usepackage{makecell}
\usepackage{colortbl}
\usepackage{minitoc}
\usepackage{tcolorbox}
\usepackage{longtable}
\usepackage{amsmath}
\usepackage{cleveref}
\usepackage{subcaption}
\usepackage{pifont}
\usepackage{framed}
\usepackage{arydshln}
\usepackage{longtable}
\definecolor{mypink}{rgb}{.99,.91,.95}
\definecolor{mygreen}{rgb}{.9,.99,.9}
\definecolor{myorange}{RGB}{255, 217, 192}
\definecolor{mygray}{gray}{.9}
\definecolor{shadecolor}{rgb}{0.9,0.9,0.9}
\definecolor{kellygreen}{rgb}{0.3, 0.73, 0.09}
\definecolor{alizarin}{rgb}{0.82, 0.1, 0.26}
\newcommand{\cmark}{{\color{kellygreen} \ding{51}}}
\newcommand{\xmark}{{\color{alizarin} \ding{55}}}
\title{DEMO: Reframing Dialogue Interaction with Fine-grained\\ Element Modeling}

\author{%
  Minzheng Wang$^{1,2}$, 
  Xinghua Zhang$^{3}$,
  Kun Chen$^{1,2}$,
  Nan Xu$^{2}$\\
  {\bf Haiyang Yu}$^{3}$,
  {\bf Fei Huang}$^{3}$,
  {\bf Wenji Mao}$^{2,1}$\thanks{Corresponding authors.}
  {\bf Yongbin Li}$^{3}$\footnotemark[1] \\
  $^{1}$ School of Artificial Intelligence, University of Chinese Academy of Sciences\\
  $^{2}$ MAIS, Institute of Automation, Chinese Academy of Sciences\\
  $^{3}$ Tongyi Lab, Alibaba Group\\
\Letter:~\texttt{\{wangminzheng2023, chenkun2024, xunan2015, wenji.mao\}@ia.ac.cn}\\ ~\texttt{\{zhangxinghua.zxh, yifei.yhy, f.huang, shuide.lyb\}@alibaba-inc.com}}

\begin{document}
\maketitle
\doparttoc
\faketableofcontents
\begin{abstract}
Large language models (LLMs) enabled dialogue systems have become one of the central modes in human-machine interaction, which bring about vast amounts of conversation logs and increasing demand for dialogue generation. The dialogue's life-cycle spans from \textit{Prelude} through \textit{Interlocution} to \textit{Epilogue}, encompassing rich dialogue elements. Despite large volumes of dialogue-related studies, there is a lack of systematic investigation into the dialogue stages to frame benchmark construction that covers comprehensive dialogue elements. This hinders the precise modeling, generation and assessment of LLMs-based dialogue systems. To bridge this gap, in this paper, we introduce a new research task—\textbf{D}ialogue \textbf{E}lement \textbf{MO}deling, including \textit{Element Awareness} and \textit{Dialogue Agent Interaction}, and propose a novel benchmark, \textbf{DEMO}, designed for a comprehensive dialogue modeling and assessment. On this basis, we further build the DEMO agent with the adept ability to model dialogue elements via imitation learning. Extensive experiments on DEMO indicate that current representative LLMs still have considerable potential for enhancement, and our DEMO agent performs well in both dialogue element modeling and out-of-domain tasks\footnote{The code and benchmark are available at \url{https://github.com/MozerWang/DEMO}}.
\end{abstract}

\begin{figure}[t]
    \centering
    \includegraphics[width=1.0\columnwidth]{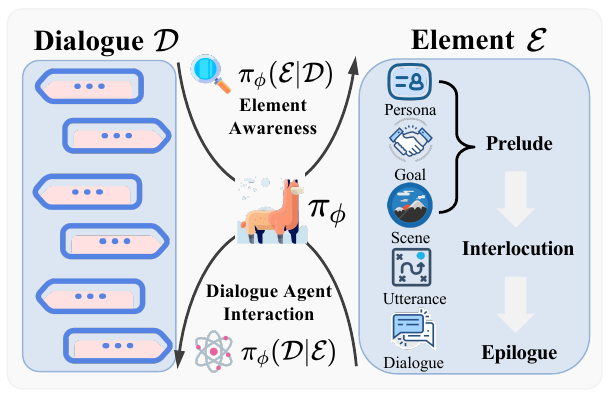}
    \caption{Overview of Dialogue Element Modeling, which focuses on two main aspects: \textit{Element Awareness} and \textit{Dialogue Agent Interaction}. We have formulated the comprehensive elements in the \textit{Prelude}, \textit{Interlocution}, and \textit{Epilogue} stages of a complete dialogue.}
    \label{fig:fig-1}
\end{figure}

\section{Introduction}
Under the compelling drive of large language models (LLMs), the development of intelligent language interfaces is undergoing an unprecedented transformation, with LLMs-empowered dialogue systems emerging as one of the central modes in human-machine interaction~\cite{ross2023programmer,sergeyuk2024ide,wuautogen}. The relentless evolution of LLMs has penetrated increasingly complex interaction environments, necessitating an enhancement of expressive intelligence~\cite{chang-chen-2024-injecting,zhou-etal-2024-characterglm,wang-etal-2024-sotopia,zhang2025sotopia} and a sharp sensitivity to the pivotal elements within interactions~\cite{tang2023enhancing,xu2024unilog}. By meticulously analyzing vast conversation logs, we can gain valuable insights into the critical elements underlying dialogue interaction, such as persona, scenario, and interaction goal, which are vital for enhancing the modeling, generation, and assessment of human-machine interaction systems. 

\begin{table*}[t]
\centering
{\setlength{\tabcolsep}{0.6pt}
\footnotesize
\scalebox{1.0}{\begin{tabular}{@{}l | c c c c c | c c c @{}}
\toprule
\textbf{Data} &  
\begin{tabular}[c]{@{}c@{}}\hspace{0.5em}\textbf{Goal}\end{tabular} & 
\begin{tabular}[c]{@{}c@{}}\hspace{0.5em}\textbf{Scene}\end{tabular} & 
\begin{tabular}[c]{@{}c@{}}\hspace{0.5em}\textbf{Persona}\end{tabular} & 
\begin{tabular}[c]
{@{}c@{}}\hspace{0.5em}\textbf{Utterance}\end{tabular} & 
\begin{tabular}[c]
{@{}c@{}}\hspace{0.5em}\textbf{Dialogue}\hspace{0.5em} \end{tabular} & 
\begin{tabular}[c]
{@{}c@{}}\hspace{0.5em}\textbf{Analysis} \end{tabular} & 
\begin{tabular}[c]
{@{}c@{}}\hspace{0.5em}\textbf{Generation} \end{tabular} & 
\hspace{0.5em} \textbf{Multilingual} \hspace{0.5em} \\
\midrule


{\textbf{DialogSum}~\citep{chen-etal-2021-dialogsum}} & \xmark & \xmark & \xmark & \xmark & \cmark & \cmark &\xmark &\xmark\\

{\textbf{SODA}~\citep{kim-etal-2023-soda}} & \xmark & \xmark & \xmark & \xmark & \cmark & \xmark &\cmark &\xmark\\

{\textbf{CharacterGLM}~\citep{zhou2023characterglm}} & \xmark & \cmark & \cmark & \xmark & \cmark & \xmark &\cmark &\xmark\\

{\textbf{Persona-Chat}~\citep{jandaghi2023faithful}} & \xmark & \xmark & \cmark & \xmark & \cmark & \xmark &\cmark &\xmark \\

{\textbf{SOTOPIA}~\citep{zhou2024sotopia}} & \cmark & \cmark & \cmark & \xmark & \xmark & \xmark &\cmark &\xmark \\

{\textbf{Ditto}~\citep{lu-etal-2024-large}} & \xmark & \xmark & \cmark & \xmark & \cmark & \xmark &\cmark &\cmark \\
\midrule
\textbf{DEMO (Ours)} & \cmark & \cmark & \cmark & \cmark & \cmark & \cmark & \cmark & \cmark\\
\bottomrule
\end{tabular}}}
\caption{The overview of our DEMO's characteristics compared to those in the related work. DEMO encompasses comprehensive dialogue elements and tasks applicable to both English and Chinese.}
\label{tab:related_benchmarks}
\end{table*}

Typically, a conversational dialogue is conducted with a goal-oriented focus, relying upon a profound understanding of its core elements~\citep{austin1975things,searle1969speech,watzlawick2011pragmatics}. \textit{Dialogue agents} navigate towards their \textit{goals} within the constraints of the \textit{scene}, utilizing conversation \textit{strategy} to interact with other \textit{participants}, ultimately producing conversational \textit{content} with communicative \textit{intent}. In reality, the dialogue's life-cycle spans from the {\it \textbf{Prelude}} through the {\it \textbf{Interlocution}} to the {\it \textbf{Epilogue}}, encompassing a variety of key elements~\citep{schegloff2007sequence, hutchby2008conversation}. However, existing dialogue benchmarks inadequately cover these comprehensive aspects~\citep{zhang2024imperative}, merely concentrating on dialogue generation within pre-established settings~\citep{liu2022go, kim-etal-2023-soda,zhou2023characterglm} or only predicting selective elements based on dialogue content~\citep{jiang-etal-2023-spm, ramprasad-etal-2024-analyzing, zhou2023learning}. Consequently, the absence of investigation into the distinct dialogue stages with these nuanced elements in the current dialogue dataset construction hinders the multifaceted modeling, understanding, and thorough evaluation of diverse dialogue-related downstream tasks.

To address the above fundamental issues, in this paper, we systematically devise a dialogue construction framework and define a new research task: \textbf{D}ialogue \textbf{E}lement \textbf{MO}deling. Concretely, our proposed task focuses on two core competencies of models: (1) \textbf{\textit{Element Awareness}}, which entails reverse engineering to decompose dialogue elements, and (2) \textbf{\textit{Dialogue Agent Interaction}}, which involves goal-directed multi-turn dialogue generation driven by fine-grained elements. Furthermore, we meticulously design a data synthesis framework for constructing a tailor-designed benchmark \textbf{DEMO}, to facilitate comprehensive dialogue modeling and assessment applicable to both English and Chinese. Besides, we amass a substantial collection of expert experiences and build a DEMO agent endowed with dialogue element modeling. We conduct extensive experiments, and the results indicate that the current advanced LLMs still have considerable space for further improvement. Our DEMO agent shows promising performances in both in-domain task as well as out-domain task for social intelligence and general tasks. The main contributions of our work are as follows:
\begin{itemize}
\item To support fine-grained dialogue analysis, generation, and assessment, we reframe the dialogue interaction process by defining a system of dialogue elements and propose a pioneering research task of dialogue element modeling.
\item To promote dialogue element modeling, we innovate an element construction framework to develop a novel, comprehensive benchmark DEMO, and craft a DEMO agent for this task.
\item Through extensive experiments, we evaluate the competencies of LLMs on DEMO, and the results show that DEMO agent performs well in both in-domain and out-of-domain tasks.
\end{itemize}

\section{Dialogue Element Modeling}
\subsection{A System of Dialogue Elements}
The dialogue is conducted with a goal-oriented focus, relying upon a deep understanding of its core elements~\citep{austin1975things,searle1969speech,watzlawick2011pragmatics}. Participants strategically navigate towards their goals within the scene's constraints, engaging with their environment to produce content with clear intent. The life-cycle of a dialogue spans from the prelude through the interlocution to the epilogue, encompassing various elements~\citep{schegloff2007sequence, hutchby2008conversation}. In the prelude, the focus is on the motivation and necessary elements of the dialogue~\citep{goffman1981forms,schiffrin1994approaches}, which include the participants' backgrounds, the time and place, the topic, and the goals of both parties. During the interlocution, attention is given to the elements intrinsic to each response, such as the intentions participants aim to convey, their current emotions and feelings, and the dialogue strategies employed~\citep{goffman1981forms,brown1987politeness}. The epilogue involves summarizing the entire dialogue, assessing the fulfillment of both parties' goals, and examining the flow of information throughout the dialogue~\citep{schegloff1973opening,drew1998figures}. For the detailed description of the framework of dialogue elements, please refer to \Cref{app:system}. 

\begin{figure*}[t]
    \centering
    \includegraphics[width=1\textwidth]{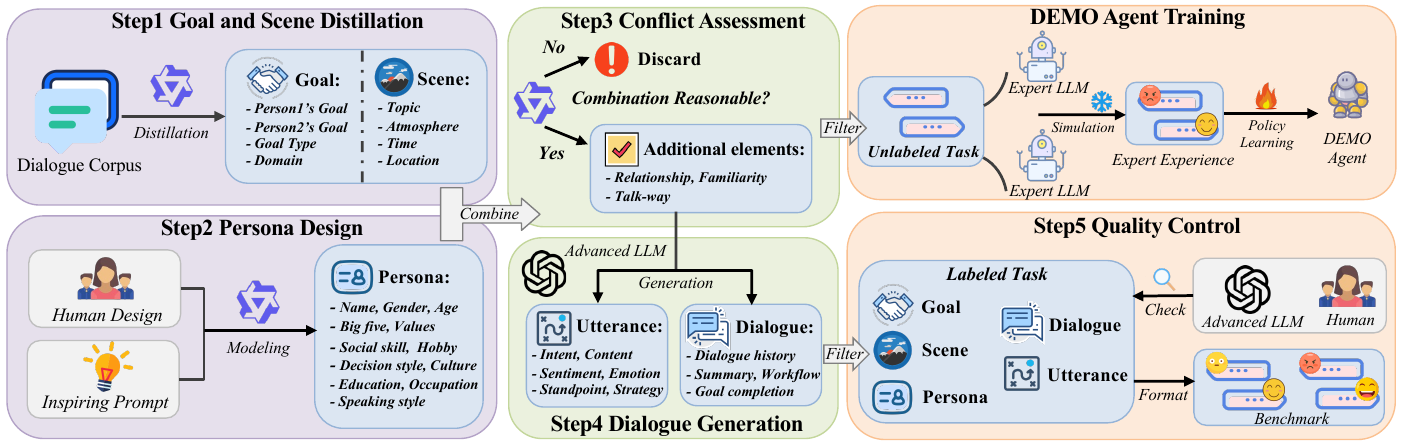}
    \caption{Overview of our DEMO synthesis framework, which consists of five steps: (1) Goal and Scene Distillation,
(2) Persona Design, (3) Conflict Assessment, (4) Dialogue Generation, (5) Quality Control. The process of DEMO Agent training is also shown in this figure.}
    \label{fig:overview}
\end{figure*}

\subsection{Task Definition}
The task of dialogue element modeling focuses on two main aspects: \textit{Element Awareness} and \textit{Dialogue Agent Interaction}. Specifically, (1) Element Awareness examines whether LLM can reverse-engineer elements such as goal, persona, and scene from the entire conversation and analyze elements at the utterance level. (2) Dialogue Agent Interaction assesses the model's goal-oriented interaction capability, evaluating whether it can achieve its goal within a given environment through a limited number of interaction rounds.

\subsubsection{Element Awareness}  
The \textit{Element Awareness} primarily focuses on offline single-turn inference. Given an entire dialogue, it aims to model the key elements that contribute to the conversation. It has four tasks: (1) \textit{Goal Recognition}, (2) \textit{Persona Modeling}, (3) \textit{Scene Reconstruction}, and (4) \textit{Utterance Mining}. 

\noindent\textbf{Goal Recognition} 
This task tends to attain the goal elements $\mathcal{G}$ from the given dialogue $\mathcal{D}$. Dialogues are arguably goal-driven~\citep{searle1969speech,austin1975things}, and this task aims to identify the behavioral motivations of participants using the model $\pi_{\phi}$. Specifically, the model needs to identify each person's dialogue goal $g_1$ and $g_2$, and the extent $s$ to which those goals are achieved. This process can be formally defined as $\pi_{\phi}(g_1,g_2,s|\mathcal{D})$. 

\noindent\textbf{Persona Modeling}
The task requires constructing the persona $\mathcal{P}$ of the two dialogue participants from the given dialogue $\mathcal{D}$. Personality, experiences, educational background, and interests often influence the manner of interaction~\citep{grice1975logic,austin1975things}, establishing a mapping relationship between persona and dialogue content. This task requires the model to infer from effect to cause, as well as reverse modeling persona from dialogue content. Specifically, based on the dialogue content $\mathcal{D}$, the model $\pi_{\phi}$ aims to infer persona $p_1$ and $p_2$, including the gender, age, personality, speaking style, hobby, and background of the two participants. This task can be formally defined as $\pi_{\phi}(p_1,p_2|\mathcal{D})$.

\noindent\textbf{Scene Reconstruction}
This task requires reasoning scene elements $\mathcal{S}$ from the given dialogue $\mathcal{D}$. The scene specifies the topic, interaction type, and the relationship and familiarity between the participants, which are crucial for making the dialogue more dynamic and nuanced~\citep{reeves1996media,pickering2004toward}. Specifically, it requires the model $\pi_{\phi}$ to reconstruct the pre-existing relationship, interaction type, and topic before the conversation starts, as well as to deduce the information flow throughout the dialogue and summarize the conversation for each participant. This task can be formalized as: $\pi_{\phi}(\mathcal{S}|\mathcal{D})$.

\noindent\textbf{Utterance Mining}
The task involves the extraction of utterance-level implicit information $\mathcal{U}$ from a given dialogue $\mathcal{D}$. Each response typically conveys rich information~\citep{goffman1981forms,brown1987politeness}, with participants employing conversational strategies, expressing standpoints and emotions, and aiming to realize their intentions for dialogue goal attainment. Specifically, given the content of a dialogue $\mathcal{D}$, the model $\pi_{\phi}$ is required to extract the intention, sentiment, emotion, stance, and strategy expressed by a particular utterance. This task can be formalized as follows: $\pi_{\phi}(\mathcal{U}|\mathcal{D})$

\subsubsection{Dialogue Agent Interaction}
The \textit{Dialogue Agent Interaction} refers to the two-party goal-directed multi-turn dialogue interaction in language space. This task encompasses a wide range of interaction types, both cooperative and non-cooperative, including persuasion, argument, empathy, negotiation, and more, which can be regarded as an incomplete information game~\citep{reif1984complexity}. It examines the ability of LLM to model dialogue driven by elements through dynamic inference. It can be formulated as a Markov Decision Process~\citep{bellman1957markovian}. 

\noindent\textbf{State} The persona, goal, scene, and dialogue history in each episode denote the state. 
As the interactions progress, the dialogue history is continuously updated while the persona, goal, and scene stay unchanged. The global state at timestep $t$ is represented as $\mathcal{D}_t$, which includes the dialogue content generated after the $t$-th turn along with other constant dialogue elements. Meanwhile, local states tied to specific sub-action sets are enhanced by combining the global state $\mathcal{D}_t$ with the history of the previous $t$-$1$ dimensional sub-action choices.

\noindent\textbf{Action} The interaction unfolds between two agents, $\mathcal{A}_1$ and $\mathcal{A}_2$. At each turn $t$, according to the observation of the state, the agent selects an action that consists of one utterance $\mathcal{U}_t$ generated by itself.

\noindent\textbf{Transition} In our setting, the transition function adds the utterance to the interaction history while the persona, goal, and scene stay unchanged in state representation. The dialogue history can be represented as an alternating sequence of utterances generated by two players, denoted by \{$\mathcal{U}_1,\mathcal{U}_2,\mathcal{U}_3,...,\mathcal{U}_t-1,\mathcal{U}_t$\}. The interaction continues until the dialogue goal is achieved or the maximum number of turns $T$ is reached.

\noindent\textbf{Reward} After each turn, a reward function can be called to quantify how well each player has been doing. The design of the reward function is critical. To comprehensively examine the LLMs' dialogue interaction capabilities, we devise a multi-dimensional reward framework, which articulates 0-10 scoring criteria for each dimension, prompting GPT-4o~\citep{openai2024gpt4} as a reward function to evaluate the interaction according to four dimensions: \textit{Goal achievement} (i.e. the extent to which the dialogue goals of both parties are fulfilled),~\textit{Believability} (i.e. the extent to how well participants align with dialogue elements),~\textit{Skillfulness} (i.e. the ability of participants to analyze dialogue history, mine utterances, and provide appropriate responses) and \textit{Realistic} (i.e. the extent to which the response content appears human-like and vivid, as opposed to being overtly AI-generated).

\section{DEMO Benchmark}
\subsection{Overview}
DEMO is our newly developed benchmark specifically designed to enhance the dialogue element modeling capabilities of dialogue systems, which features an equal 1:1 ratio of Chinese to English languages. To evaluate element awareness, we have a total of 4,000 evaluation samples, with a maximum of 26 dialogue turns and an average of 18.3 turns, covering 23 distinct dialogue elements. Each of the four tasks contains 1,000 test samples. In terms of dialogue agent interaction, DEMO provides 1,000 episodes that cover a wide array of cooperative and non-cooperative interaction types. Examples of specific tasks are illustrated in \Cref{app:test-case}.

\subsection{Benchmark Construction Framework}
Our framework is depicted in \Cref{fig:overview}. We follow the dialogue element system to sequentially annotate the elements of prelude, interlocution, and epilogue. First, based on \textit{Goal and Scene Distillation} and \textit{Persona Design}, we generate the three prelude elements: goal, scene, and persona. Then, through the \textit{Conflict Assessment}, we ensure that the combined prelude elements are reasonable. Subsequently, we generate the corresponding interlocution and epilogue elements through \textit{Dialogue Generation}. Finally, after \textit{Quality Control}, we inspect and revise the data to establish the final benchmark.

\noindent\textbf{Goal and Scene Distillation} In this initial stage, we distillate the goals and scenes from the given dialogue. By leveraging an instance-driven paradigm, we diversify data from the large-scale dialogue corpus. We utilize SODA~\citep{kim-etal-2023-soda} and LCCC~\citep{wang2020chinese} as our seed corpora, which include millions of English and Chinese dialogues encompassing various aspects of social commonsense. Specifically, we employ Qwen2-72B~\citep{yang2024qwen2} to extract participant's goals and the conversation scene from dialogue. Finally, we get 2.6 Million goal and scene data.

\noindent\textbf{Persona Design}
In parallel with the previous phase, we established a comprehensive and diverse persona collection. The creation of personas is divided into two parts: (1) Designing the persona attributes pool and (2) Inspiring prompting. Referring to ~\citep{zhou2023characterglm, zhou2024sotopia, chen2024persona, yang2024simschat}, we consider the following attributes: name, gender, age, Big Five traits ~\citep{mccrae1992introduction}, moral values~\citep{graham2011mapping}, social skills~\citep{yang2024social}, personal values~\citep{schwartz1992universals}, and decision style ~\citep{scott1995decision}. Based on these characteristics, we combine them and then leverage Qwen2-72B for more detailed persona modeling. By prompting the LLM with diverse web texts, as~\citep{chan2024scaling} suggests, we generate wide-ranging personas. Ultimately, the LLM produces detailed information on each person’s background, hobbies, education, occupation, culture, relationships, and speaking style. At this stage, we have modeled 200,000 diverse personas.

\noindent\textbf{Conflict Assessment}
\label{method-ca}
After gathering prelude elements, we proceed to acquire reasonable combinations of these elements. To assess the reasonableness and coherence of the combined prelude elements, we prompt the Qwen2-72B to check for issues like character identity contradictions, misalignment between persona and goal, or unsuitable pairings of dialogue participants. For combinations deemed reasonable, we further instruct the model to provide additional details about the relationship and familiarity between participants, as well as mode of interaction. To further evaluate the LLM's capability in conflict assessment, we also conducted a human evaluation, where two human raters and the LLM performed conflict checks on the combinations following identical procedures and criteria. The inter-rater agreement measured by Cohen's Kappa~\citep{fleiss1971measuring} between the two human raters was 0.85, while the Kappa value between Rater1 and LLM is 0.65, and the value between Rater2 and LLM is 0.79. The results demonstrate high consistency between LLM and human annotators in this task, indicating that LLM performs at a human-comparable level and is competent for this annotation task. For the complete human annotation process and information, please refer to \Cref{app:human annotation}.

\begin{table}[t!]
\centering
\small
\resizebox{\columnwidth}{!}{%
\begin{tabular}{lccccc}
\toprule
 \textbf{Human} &  \textbf{Goa.} &\textbf{Per.} &\textbf{Sce.} &\textbf{Utt.} &\textbf{Avg.}  \\
\midrule
Rater1 &  0.80 & 0.87 & 0.71 & 0.74 & 0.82\\
Rater2 &  0.74 & 0.76 & 0.77 & 0.79 & 0.78\\

\bottomrule
\end{tabular}
}
\caption{The Kappa consistency results between LLM annotations and two human raters on different elements}
\label{tab:human-ana}
\end{table}
\noindent\textbf{Dialogue Generation}
Upon establishing reasonable combinations, we proceed to generate interlocution and epilogue elements. To manage data distribution effectively, we categorize all combinations into ten types based on the dialogue goal type and extract them evenly. Leveraging the LLM's role-play capabilities ~\citep{zhou2023characterglm, zhou2024sotopia, chen2024persona}, we prompt it to create dialogues that align with specified persona, goal, and scene. We also prompt the advanced LLMs to analyze each utterance with its associated intention, sentiment, emotion, stance, and strategy, culminating in a comprehensive output that includes the information flow and dialogue summary. To form our benchmark, we curate 1,800 distinct combinations, utilizing GPT-4o for generation.

\definecolor{deepgreen}{RGB}{0, 70, 0}
\definecolor{deepred}{RGB}{255, 0, 0}
\definecolor{backgreen}{RGB}{226, 240, 217}
\newcommand{\highg}{\cellcolor{backgreen}}

\begin{table*}[t]
\centering
\resizebox{1\textwidth}{!}{%
\begin{tabular}{lccccccccccc}
\toprule
\multirow{2}{*}{\textbf{Model}}   & \multicolumn{5}{c}{$\textbf{Element Awareness}$} & \multicolumn{5}{c}{\textbf{Dialogue Agent Interaction}}      & \multirow{2}{*}{\textbf{Overall}} 

\\
\cmidrule(lr){2-6} \cmidrule(lr){7-11}
      & \multicolumn{1}{c}{$Goa.$}
      & \multicolumn{1}{c}{$Per.$}
      & \multicolumn{1}{c}{$Sce.$}
      & \multicolumn{1}{c}{$Utt.$}
      & \multicolumn{1}{c}{\textbf{Avg}}
      & \multicolumn{1}{c}{$Goa.$}
      & \multicolumn{1}{c}{$Bel.$}
      & \multicolumn{1}{c}{$Ski.$}
      & \multicolumn{1}{c}{$Rea.$}
      & \multicolumn{1}{c}{\textbf{Avg}} \\
\midrule

\multicolumn{12}{l}{\textit{Proprietary LLM}} \\
GPT-4o  &\underline{5.975}	&4.051	&6.167	&7.308	&5.875	&\underline{8.190}	&\underline{9.181}	&\underline{8.614}	&\underline{8.537}	&\textbf{8.631}	&\textbf{6.793}\\
Claude-3.5-Sonnet  &\textbf{5.979}	&\underline{4.145}	&6.221	&6.243	&5.647	&7.571	&9.174	&8.432	&\textbf{8.840}	&\underline{8.504}	&6.599\\
GPT-4o-mini &5.802	&3.586	&5.748	&7.002	&5.534	&7.551	&9.082	&8.316	&8.163	&8.278	&6.449 \\
Claude-3.5-Haiku &5.492	&3.858	&6.071	&6.304	&5.431	&7.361	&9.115	&8.402	&7.968	&8.212	&6.358\\
\hdashline

\multicolumn{12}{l}{\textit{Open-sourced LLM}} \\
Qwen2-72B-Instruct  &5.357	&\textbf{4.406}	&5.702	&6.921	&5.596 &\textbf{8.447} &\textbf{9.204} &\textbf{8.699} &8.175 &\textbf{8.631} &6.608\\ 
Llama-3.1-70B-Instruct &5.559	&3.643	&6.078	&7.051	&5.593	&7.223	&7.914	&7.222	&6.753	&7.278	&6.154  \\

\midrule

\multicolumn{12}{l}{\textit{Backbone LLM}} \\

Qwen2-7B-Instruct &5.306 &3.981	&5.459	&6.347	&5.244	&6.698	&8.112	&6.895	&6.278	&6.996	&5.828\\

Llama3.1-8B-Instruct &5.546	&3.287	&5.403	&6.523	&5.189	&5.831	&6.166	&5.519	&4.974	&5.623	&5.335 \\

\hdashline

\multicolumn{9}{l}{\textit{DEMO Agent}} \\
DEMO-Qwen2-7B &5.229	&3.946	&\underline{6.534}	&\underline{7.914}	&\underline{5.906}	&7.450	&8.864	&8.073	&7.864	&\text{8.063}	&\underline{6.625}\\
& -& -&$\Delta 1.075$ &$\Delta 1.567$ &$\Delta 0.797$& $\Delta 0.752$& $\Delta 0.752$ & $\Delta 1.178$ &$\Delta 1.586$ &$\Delta 1.067$ &$\Delta 0.797$\\

DEMO-Llama3.1-8B &5.623	&3.939	&\textbf{6.543}	&\textbf{7.926}	&\textbf{6.008} &6.945	&7.688	&7.015	&6.378	&\text{7.707} &\text{6.341}\\
&$\Delta 0.077$ &$\Delta 0.652$ &$\Delta 0.752$ &$\Delta 1.140$ & $\Delta 0.819$ &$\Delta 1.114$ &$\Delta 1.522$ &$\Delta 1.496$& $\Delta 1.404$& $\Delta 2.084$ &$\Delta 1.006$ \\

\bottomrule
\end{tabular}
}
\caption{The results of various LLMs on the DEMO. The highest score among different LLMs is highlighted in \textbf{bold}, and the second highest is \underline{underlined}. And $\Delta$ values represent the improvement over the baseline. \textbf{Element Awareness} has four tasks:(1)~\textit{Goa}: Goal Recognition, (2)~\textit{Per}: Persona Modeling, (3)~\textit{Sce}: Scene
Reconstruction, and (4)~\textit{Utt}: Utterance Mining. \textbf{Dialogue Agent Interaction} includes four dimensions: (1)~\textit{Goa}: Goal Achievement, (2)~\textit{Bel}: Believability,(3)~\textit{Ski}: Skillfulness,(4)~\textit{Rea}: Realistic. \textbf{Overall} is the average score of two tasks.}
\label{tab:main-res}
\end{table*}
\noindent\textbf{Quality Control}
\label{paper:quality-control}
To ensure the accuracy of benchmark annotations, we employ a three-step verification process: (1) Advanced-LLM Check: Two of the most advanced LLMs, GPT-4o and Claude-3.5-Sonnet, independently review and validate the quality of annotations. They examine each entry, editing any unreasonable or low-quality labels to maintain accuracy. (2) Voting: We implement the simple majority voting method to finalize the label. (3) Manual Check: After the voting process, We engage two human raters to further examine and assess the quality of data annotation following our pre-established quality control standards. For the detailed human annotation information, please refer to \Cref{app:human annotation}. 

We also use Kappa score~\citep{fleiss1971measuring} to measure annotation quality in \Cref{tab:human-ana}. The data quality inspection process is performed by two experienced annotators, with a consistency Kappa value of 0.84 between them. Their Kappa consistency results with different elements annotated by the LLM are shown in \Cref{tab:human-ana}. The LLM shows high consistency with the two annotators, demonstrating performance comparable to humans. Additionally, we conducted a manual verification of the data, achieving an accuracy rate of 91.17\%. These results have all validated the quality of our benchmark.

\subsection{DEMO Agent Training}
To further investigate the task characteristics and impact of fine-grained elements, we build the DEMO agent endowed with dialogue element modeling. Humans have the ability to learn efficiently through observing and imitating the behavior of others \citep{NIPS1996_68d13cf2, pmlr-v15-ross11a, torabi2018behavioral}. Behavioral learning~\citep{10.5555/647636.733043, ross2010efficient}, is an approach to extracting and distilling expert policies from high-quality data, particularly from models with advanced capabilities. Drawing inspiration from these, we propose enhancing the performance of LLMs in our task by developing an imitation policy. 
In the context of dialogue element modeling, it involves analyzing our pre-defined elements in the Prelude, the Interlocution, and the Epilogue through single-turn reasoning, as well as modeling multi-turn dialogue content based on comprehensive fine-grained dialogue elements.
During the benchmark construction process, we have accumulated a diverse amount of unlabeled data, which serves as the environment for the expert model to simulate dialogue modeling. Specifically, GPT-4o is employed as the expert model. By engaging in both single-turn and multi-turn interactions within this environment, we are able to amass a wealth of expert experience, which is subsequently utilized to train the model.

\section{Experiments}
\subsection{Experimental Setup}
\textbf{Models} We evaluate ten advanced LLMs, including API-based LLMs: GPT-4o, GPT-4o-mini, Claude3.5-Sonnet~\citep{anthropic2024claude35sonnet}, Claude3.5-Haiku~\citep{anthropic2024claude35haiku} and Open-sourced LLMs: Qwen2-72B-Instruct, Qwen2-7B-Instruct, Llama3.1-70B-Instruct, Llama3.1-8B-Instruct~\citep{dubey2024llama}.

\noindent\textbf{Evaluation Metric} Evaluating the unpredictable behaviors of LLMs, traditional metrics such as BLEU and Rouge-L may yield inaccurate responses. Recent research~\citep{zhang2023wider, zheng2023judging, kim2023prometheus} indicates that the GPT-4 evaluator demonstrates high consistency with human evaluation while reducing costs, making it a reasonably reliable annotator. Following these work~\citep{perez-etal-2022-red, zhou2024sotopia, wang-etal-2024-leave}, we prompt GPT-4o as a judge model. For the element awareness task, we evaluate the output based on the golden answer from several aspects, scoring from 0 to 10. For dialogue agent interaction task, we only consider the reward at the end of the interaction. The detailed prompts are provided in \Cref{app:GPT4-as-the-Judge}. We also conduct human evaluation on LLM evaluator in \Cref{app:human-llm-judge}.

\noindent \textbf{Implement Details} To ensure the stability of the evaluation, we set the temperature of the evaluator to 0. For element awareness, we set the temperature to 0 to ensure reproducibility. For dialogue agent interaction, we set the temperature to 1 to encourage diversity. Please refer to \Cref{app:Model Versions} for more details about training and the used LLMs.

\subsection{Main Results}
\begin{table*}[t!]
\centering
\small
\resizebox{1\textwidth}{!}{%
\begin{tabular}{lcccccccc}
\toprule
 \textbf{Model} &  \textbf{SOC} &\textbf{SEC} &  \textbf{FIN} &\textbf{REL}&  \textbf{KNO} &\textbf{GOA}&  \textbf{BEL} &\textbf{Overall} \\
\midrule

Qwen2-7B-Instruct & -0.05 &0.00 & 0.73 &1.83 &3.41 &6.07 &8.64 &2.95 \\

DEMO-Qwen2-7B & -0.02 &0.00 &0.82 &2.32 &4.52 & 6.40 &8.94 &3.28$(\Delta 0.33)$\\

\hdashline

Llama3.1-8B-Instruct &  -0.50 &-0.01 &-0.16 &-0.60 &2.21 &3.39 &8.63 &1.85\\

DEMO-Llama3.1-8B &  -0.19 &0.00 &0.29 &0.85 &2.88 &3.77 &8.41 &2.29 $(\Delta 0.44)$\\

\bottomrule
\end{tabular}
}
\caption{Evaluation results on SOTOPIA, which scored from seven social dimensions
. The overall score is the average of the seven dimensions reflecting the overall social intelligence. GPT-4o rates each dimension.}
\label{tab:sotopia-eval}
\end{table*}
We assess ten advanced LLMs on the DEMO benchmark. The main results are shown in \Cref{tab:main-res}. To cross-validate results, we add two other exemplar LLMs, DeepSeek-V3~\citep{liu2024deepseek} and Gemini-1.5-Pro~\citep{team2024gemini}, as judge models (see \Cref{app:two-judge-model}).

\noindent\textbf{Model Analysis} GPT-4o shows the best overall performance, maintaining great performance across all dimensions. Analyzing from the perspective of parameter size, the model's performance aligns with the \textit{Scaling Law}, indicating that models with larger parameters possess stronger expressive capabilities. Additionally, the gap between open-source and closed-source models is narrowing. For instance, Qwen2-72B-Instruct has achieved state-of-the-art performance in dialogue agent interaction tasks, with overall performance differences from GPT-4o being minimal.

\noindent\textbf{Task Analysis} There remains significant room for improvement in dialogue element modeling, particularly in element awareness task. Accurately modeling various elements (such as persona modeling) from dialogue content is still challenging, potentially requiring multi-step reasoning or additional clues. In dialogue agent interaction tasks, current LLMs exhibit excellent expressive capabilities, adeptly adhering to settings and generating relatively realistic content. Humans are inherently social, striving to achieve social goals in daily interactions. Goal achievement is a crucial feature of intelligence; thus, the ability to perceive targets and collaborate to achieve goals reflects LLMs' higher-order capabilities. However, their ability to achieve self-set goals through multi-turn interactions requires enhancement.

\noindent\textbf{DEMO Agent} By learning through expert experience imitation, the DEMO agent has achieved significant improvements across two different backbones, with an average task improvement of 0.9. Specifically, the agent utilizing the LLaMA backbone achieved SOTA performance in element awareness tasks. Meanwhile, the agent built on the Qwen backbone secured the second-highest score, surpassed only by GPT-4o. The DEMO Agent has also surpassed or performed on par with models with larger parameters, such as Claude3.5-Sonnet and the Qwen2-72B-Instruct. This demonstrates the effectiveness of imitation learning and expert experience. However, this method has a performance ceiling limited by the abilities of the expert model. Fully leveraging additional modeling cues to develop the capacity between element awareness and intelligent interaction will be a primary focus of our future work. We also present the detailed case study to analyze LLM outputs in \Cref{app:case-study}.

\subsection{Out-of-domain Performance}
The DEMO agent has demonstrated promising results in dialogue element modeling within the domain. However, the question remains: can this capability extend to tasks beyond that domain? To evaluate this, we selected the hard episodes of SOTOPIA~\citep{zhou2024sotopia} as our testing environment, which assesses social intelligence. Two LLMs are prompted to act as role-playing social agents in SOTOPIA, engaging with each other through communication. SOTOPIA designed a seven-dimension framework to assess the social intelligence of LLMs: social rules, secret-keeping, financial benefits, relationship maintenance, knowledge, goal completion, and believability. For each task, agents are scored along designed dimension.

\Cref{tab:sotopia-eval} presents the results. All our DEMO agents show remarkable generalization capabilities in social intelligence tasks, with prominent performance improvements. This validates the necessity and effectiveness of fine-grained dialogue modeling. For baseline models like Llama3.1-8B-Instruct, during pre- and post-training phases, the training data only included dialogue content or very limited dialogue elements. Dialogue elements are implicitly learned from the conversation content, thus resulting in inferior overall performance. In contrast, the DEMO agent, through explicitly modeling and learning various elements of a dialogue, can better understand dialogue content, clearly recognize current goals and intentions, and follow the current dialogue scene and persona settings, thereby performing more competently in complex social interaction scenarios.

\subsection{Catastrophic Forgetting Problem}
In addition to confirming the model's great performance in dialogue element modeling, evaluating whether the other capabilities remain unaffected is equally crucial. Continued training can sometimes lead to catastrophic forgetting, where the model loses previously acquired knowledge, disrupting its initial alignment. We use the Helpful, Honest, Harmless (HHH)~\citep{askell2021general} dataset to assess the impact on alignment performance. This involves a multiple-choice task to measure the model's ability to select better answers from two given options. When presented with both options, we calculate the model's tendency to favor one answer over the other. To assess the model's general capabilities, we employ the MMLU~\citep{hendrycks2021measuring}, using a 5-shot evaluation based on next-word prediction. Accuracy serves as the evaluation metric across two benchmarks. 

The results are presented in \Cref{tab:mmlu-eval}. It is evident that DEMO Agents retain the overall capabilities of the base model. Although a few did not exhibit enhancements, our models performed comparably to the base model. They did not experience significant issues with catastrophic forgetting, indicating that the dialogue element modeling operates independently of the general capabilities.
\begin{table}[t!]
\centering
\small
\resizebox{\columnwidth}{!}{%
\begin{tabular}{lcc}
\toprule
 \textbf{Model} &  \textbf{MMLU} &\textbf{HHH}  \\
\midrule
Qwen2-7B-Instruct &  69.04 &45.70\\

DEMO-Qwen2-7B &68.37 &46.15$(\Delta 0.45)$\\

\hdashline

Llama3.1-8B-Instruct &  65.94 &46.61\\

DEMO-Llama3.1-8B &66.06 $(\Delta 0.12)$&45.25\\

\bottomrule
\end{tabular}
}
\caption{The results of the LLM's general capability and alignment performance, using the accuracy score.}
\label{tab:mmlu-eval}
\end{table}
\section{Related Work}
We review related research based on two trends in interactive dialogue systems: (1) \textit{Dialogue Analysis}: inferring dialogue elements based on dialogue content. (2) \textit{Dialogue Generation}: generating dialogue content according to the dialogue settings. While we discuss related work from the perspectives of two trends mentioned above, different types of dialogue systems and their distinction are discussed in \Cref{app:dialogue-type}.

\subsection{Dialogue Analysis}
The goal of dialogue analysis is to mine critical elements (such as intent, profiles, summary, etc.) from the dialogue~\citep{zhang2024imperative}, which can extract actionable insights and drive empowerment. In the era of small language models, dialogue analysis did not form a systematic task but was broken down into atomic tasks, such as slot filling and intent classification~\citep{qin-etal-2020-agif, louvan2020recent, jiang-etal-2023-spm}, state tracking~\citep{eric-etal-2020-multiwoz,feng-etal-2023-towards,niu-etal-2024-enhancing}, dialogue summary~\citep{chen-etal-2021-dialogsum, fabbri2021convosumm, ouyang-etal-2023-compositional, ramprasad-etal-2024-analyzing} and persona extraction~\citep{wang-etal-2022-extracting, zhou2023learning}, etc. These atomic tasks are primarily concerned with task-relevant states, and they do not address the fundamental elements essential for dialogue understanding and modeling. In the era of LLM, recent work~\citep{zhang2024imperative} performed a thorough review and systematized conversation analysis task. There is a scarcity of datasets that encompass all essential elements of dialogue, which hinders the development of diverse dialogue downstream tasks.

\subsection{Dialogue Generation}
The related work on dialogue generation primarily focuses on the relevant indicators of generated content, but rarely pays attention to the process of dialogue modeling. Task-oriented dialogue~\citep{rashkin-etal-2019-towards, Sun_Moon_Crook_Roller_Silvert_Liu_Wang_Liu_Cho_Cardie_2021, liu2022go} focuses on completing specific tasks, emphasizing task completion rather than generalization. Open-domain dialogue~\citep{li-etal-2017-dailydialog,wang2020chinese,kim-etal-2023-soda} is mainly designed for "chit-chat" between users, with more general tasks and a greater focus on immersion. Recently, several role-playing works~\citep{zhou2023characterglm,lu-etal-2024-large,chen2024persona,yang2024simschat,zhou2024sotopia,zhou2024real} have emerged, which place more emphasis on dialogue engagement and character consistency. However, there is limited guidance for dialogue modeling, and they do not provide a systematic framework for modeling the full spectrum of dialogue elements. To address this, we define a more comprehensive dialogue generation task space, considering more dialogue modeling elements. We evaluate dialogue interaction capabilities through fine-grained elements and multi-turn interactions. Additionally, we introduce element awareness tasks to thoroughly assess the LLM in both Chinese and English.

\section{Conclusion}
In this work, we redefine the dialogue framework and introduce a new research task: Dialogue Element Modeling. This task involves two fundamental capabilities, element awareness and dialogue agent interaction, which enhance the complex modeling and comprehensive evaluation of dialogue systems. We first develop a process for constructing dialogue elements and create the benchmark, DEMO, which includes comprehensive dialogue elements suitable for both English and Chinese. We further develop the DEMO agent specifically for dialogue element modeling. Through extensive experiments, we assess the performance of several advanced LLMs, revealing that there is still room for improvement in this task. In addition, the results demonstrate that our DEMO agent delivers excellent performance in both in-domain and out-of-domain scenarios.

\section{Limitations and Future Work}
First, to ensure the quality of data annotation for benchmark construction, we employ two human raters to assess the conflicts and accuracy of the constructed benchmark. This process is both time-consuming and costly. 
Second, the performance of our DEMO agent is constrained by the capabilities of the expert model we used. The interplay between element awareness and dialogue agent interaction remains insufficiently understood. 
Third, our current training paradigm lacks a heuristic joint training design that could bridge the analysis and generation phases; a joint learning approach could enhance both dialogue agent interaction and element awareness.

In this work, we focus on developing the dialogue element system. We employ synthetic generation to ensure that the dialogue closely aligns with the predefined elements. In our future work, we will use real dialogue and focus on identifying which elements contribute most significantly to creating natural and realistic dialogue. Furthermore, we plan to leverage recent advances in reasoning methodologies, particularly the Long-CoT paradigm, which has demonstrated significant improvements in handling complex tasks~\citep{jaech2024openai,o3mini,team2024qwq,gemini2flashthinking,guo2025deepseek}.
Building on the pre-established work in enhancing Long-CoT reasoning through RL post-training~\citep{shao2024deepseekmath,sheng2024hybridflow,guo2025deepseek,muennighoff2025s1,wang2025ampo,yu2025dapo}, we aim to apply these approaches to dialogue element modeling. This integration could improve element analysis accuracy through test-time scaling and enable more comprehensive modeling of multi-turn dialogues by allowing for deeper, more thorough reasoning processes.
\section*{Acknowledgments}
This work is supported in part by the National Natural Science Foundation of China under Grants \#72293575, \#72225011, \#72434005 and \#62206287, and Beijing Municipal Science \& Technology Commission, Administrative  Commission of Zhongguancun Science Park under Grant \#Z231100007423016.
\bibliography{custom}

\appendix
\section{Human Annotation Process}
\label{app:human annotation}
\subsection{Rater Information}
The annotators were hired and not co-authors. This research was collaborated with the company, who provided GPU and data annotations. The annotations were conducted by 2 Ph.D. students (candidates after one and half years of study in our country typically) working as interns in this company. The annotators specialize in dialogue systems and natural language processing. They are native Chinese speakers with proficient English reading and writing skills. They worked for two weeks, following an 8-hour workday schedule, with a daily internship compensation of 500 CNY per day. Before the annotation process, we conducted specific training sessions, providing detailed annotation examples and requirements. Regular meetings were held to discuss issues encountered during the annotation process to ensure accurate data labeling. During the annotation process, the two annotators worked independently. 

\begin{figure*}[t]
    \centering
    \begin{subfigure}[b]{1\textwidth}
        \centering
        \includegraphics[width=1\textwidth]{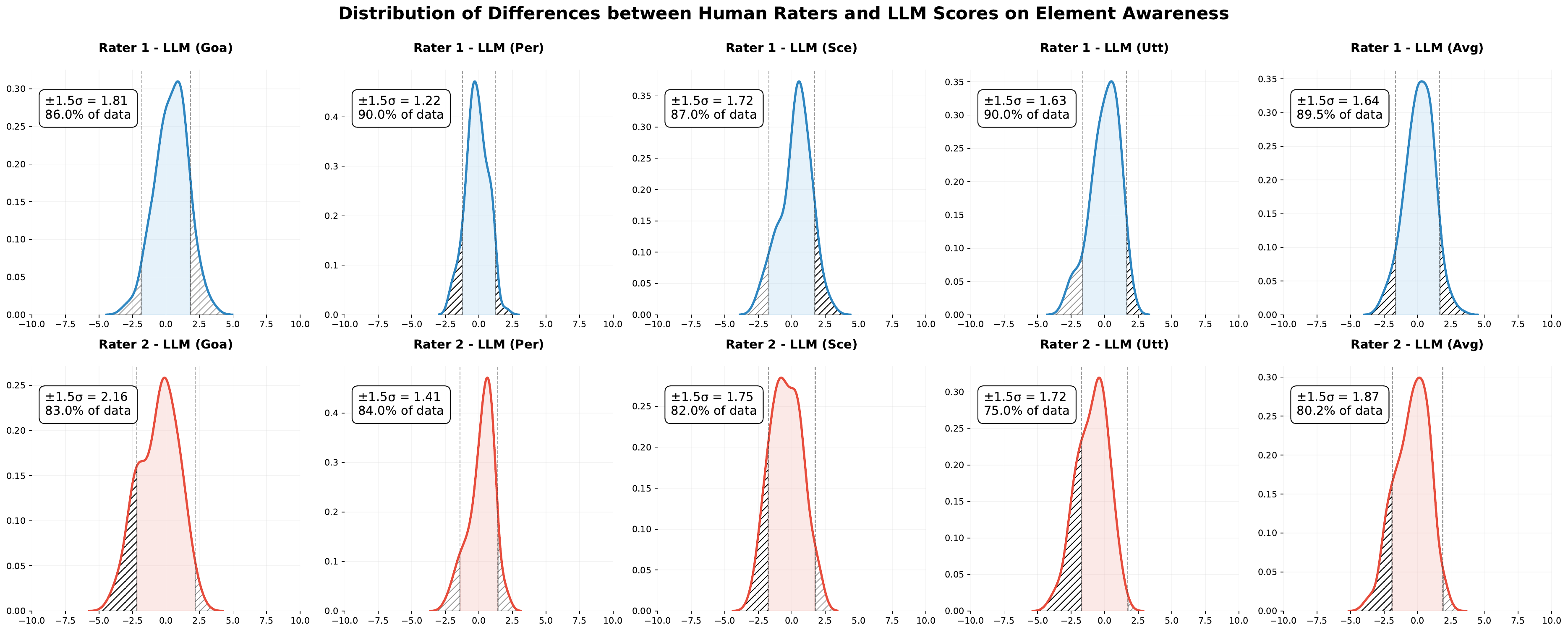}
        \caption{Element Awareness}
        \label{fig:ea-distribution}
    \end{subfigure}
    \begin{subfigure}[b]{1\textwidth}
        \centering
        \includegraphics[width=1\textwidth]{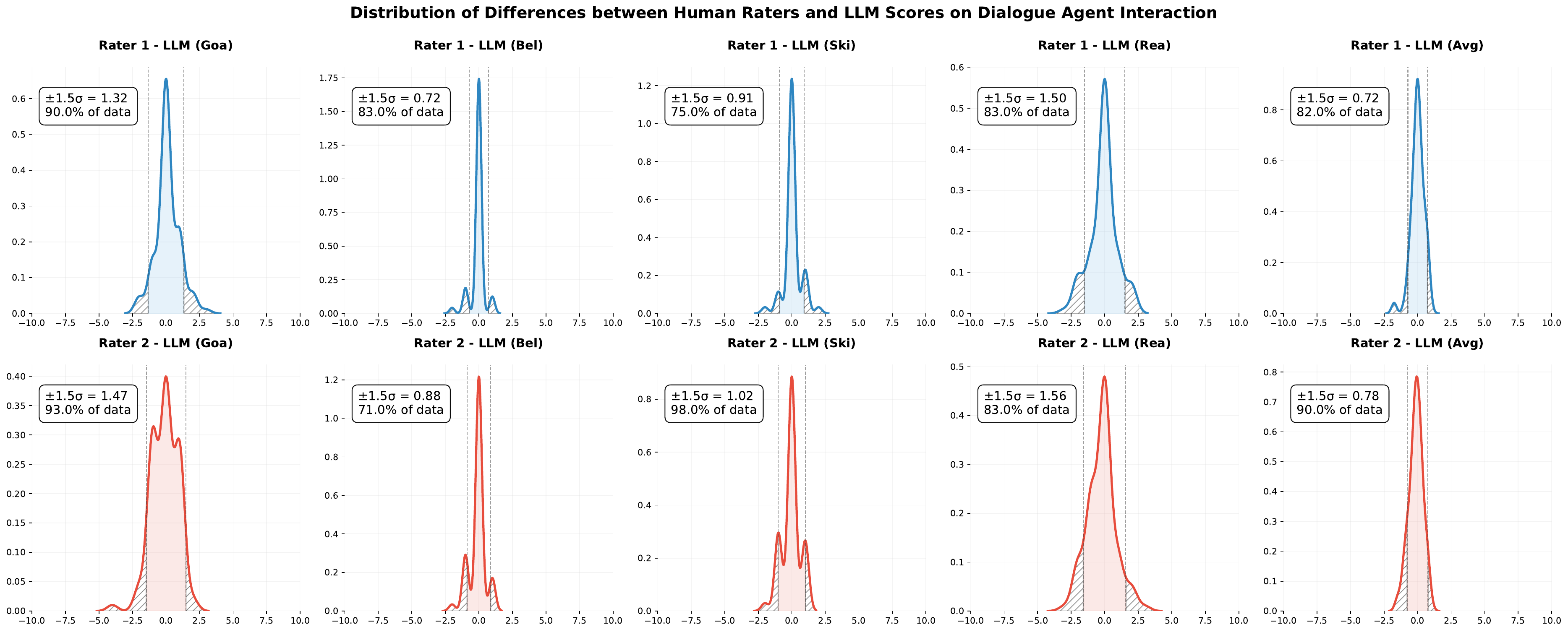}
        \caption{Dialogue Agent Interaction}
        \label{fig:dai-distribution}
    \end{subfigure}
    \caption{Distribution of score differences between human and GPT-4o evaluations on (a) Element Awareness and (b) Dialogue Agent Interaction. The x-axis shows the human-LLM score, and the y-axis shows the probability density.}
    \label{fig:distributions}
\end{figure*}

\subsection{Guidelines for Human Evaluation}
\begin{figure*}[t]
    \centering
    \includegraphics[width=0.6\textwidth]{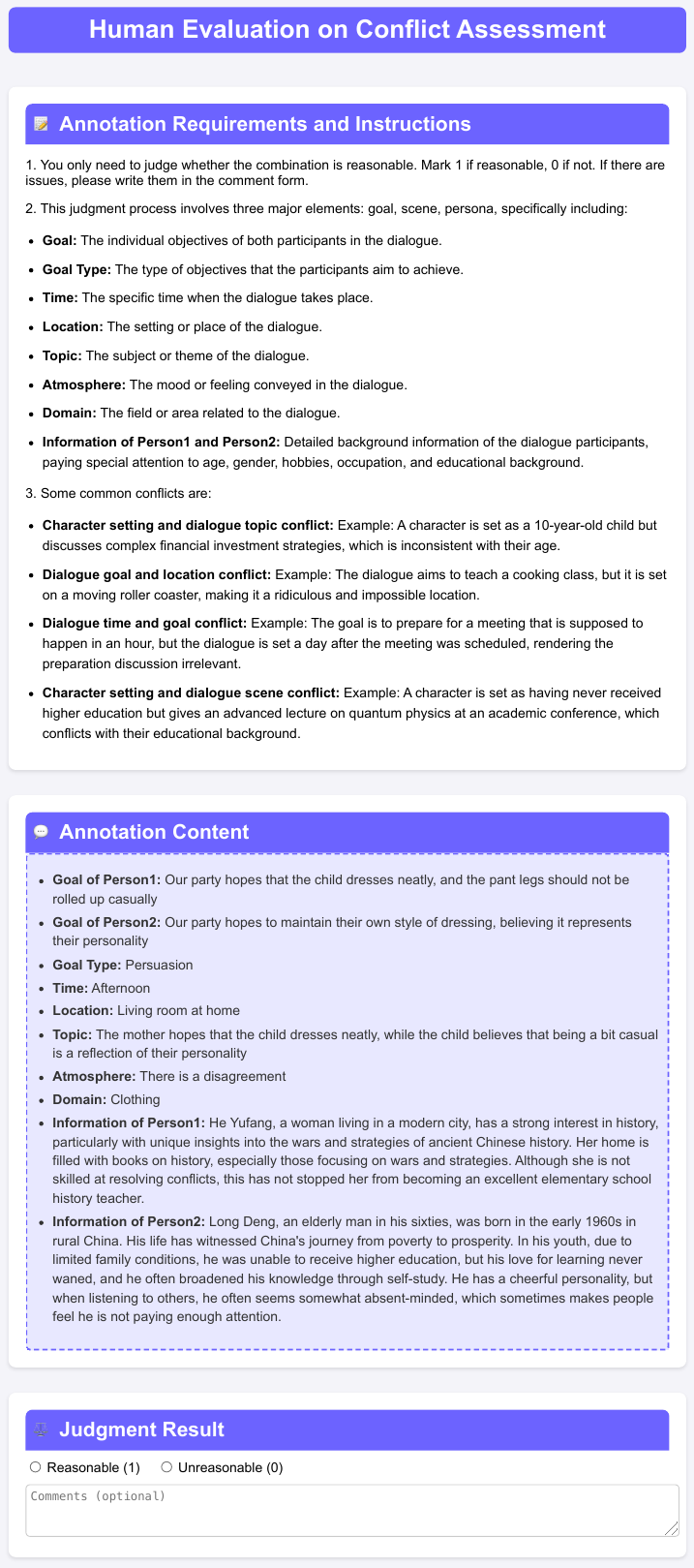}
    \caption{Guidelines and Interface for Human Annotation in Conflict Assessment}
    \label{fig:ca-check}
\end{figure*}

\begin{figure*}[t]
    \centering
    \includegraphics[width=0.5\textwidth]{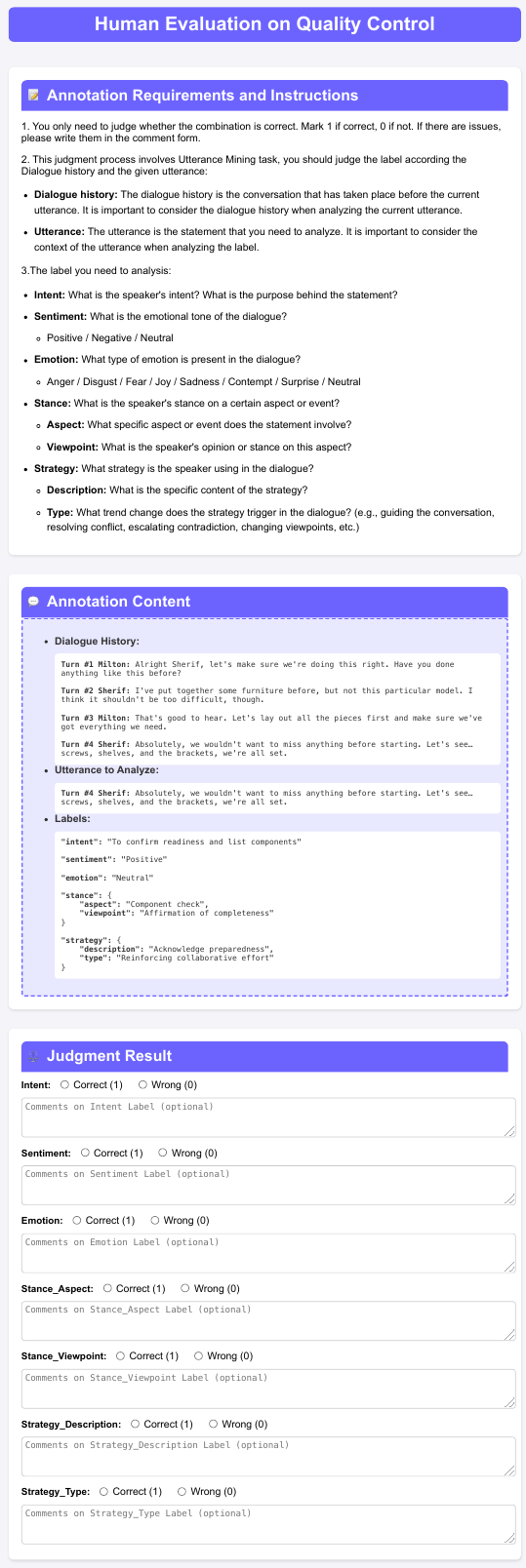}
    \caption{Guidelines and Interface for Human Annotation in Quality Control}
    \label{fig:qc-check}
\end{figure*}

The detailed annotation guidelines and interface are shown in \textbf{\textit{\Cref{fig:ca-check}}} and \textbf{\textit{\Cref{fig:qc-check}}}. The manual annotation process comprises two primary phases: Conflict Assessment and Quality Control. 

\begin{table}[t!]
\centering
\small
\resizebox{\columnwidth}{!}{%
\begin{tabular}{lcccc}
\toprule
\multicolumn{5}{c}{\textbf{\textit{Element Awareness}}} \\
\midrule
\textbf{Human} &  \textbf{Goa.} &\textbf{Per.} &\textbf{Sce.} &\textbf{Utt.} \\
\midrule
Rater1 & $0.87$ & $0.74$ & $0.78$ & $0.78$ \\
Rater2 & $0.82$ & $0.66$ & $0.78$ & $0.76$ \\
\midrule
\multicolumn{5}{c}{\textbf{\textit{Dialogue Agent Interaction}}} \\
\midrule
\textbf{Human} &  \textbf{Goa.} &\textbf{Bel.} &\textbf{Ski.} &\textbf{Rea.} \\
\midrule
Rater1 & $0.89$ & $0.64$ & $0.78$ & $0.63$ \\
Rater2 & $0.88$ & $0.42$ & $0.75$ & $0.54$ \\
\bottomrule
\end{tabular}
}
\caption{Pearson correlation coefficients between human judgment and GPT-4o evaluation on models' output among different dimensions.}
\label{tab:pearson-ea}
\end{table}

\noindent\textbf{Conflict Assessment} In this phase, we aim to evaluate the agreement between human raters and the LLM in determining the reasonableness of combinations. This helps assess the model's judgment capabilities and its suitability for this annotation task. We sample 300 unlabeled combinations of elements (scene, goal, and persona), each independently reviewed by humans and LLM. Annotators are tasked with evaluating whether each combination is reasonable and conflict-free, assigning a binary score: 1 for reasonable combinations and 0 for unreasonable ones. The results, as shown in \textbf{\textit{\Cref{method-ca}}}, indicate a high level of consistency between LLM and human assessments, thereby validating the quality of the annotations.

\begin{figure*}[t]
    \centering
    \includegraphics[width=0.9\textwidth]{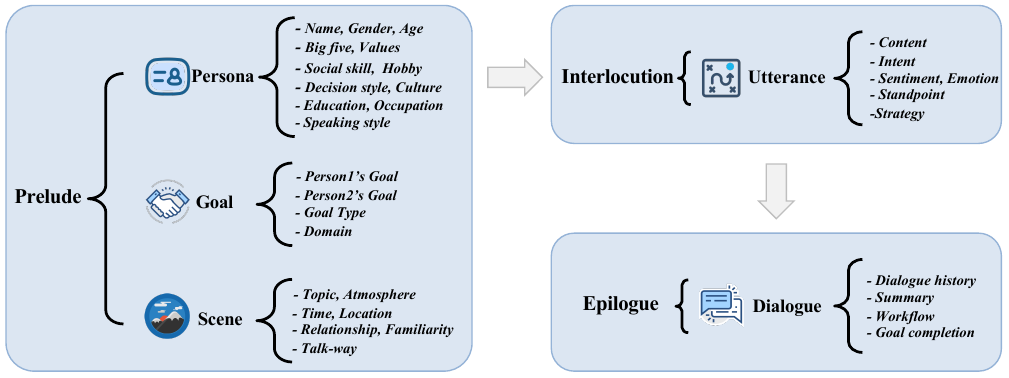}
    \caption{Overview of the system of dialogue elements.}
    \label{fig:overview-system}
\end{figure*}

\noindent\textbf{Quality Control} To assess the agreement between humans and the LLM in quality control and error detection, both human annotators and the LLM independently review 1,200 unchecked data points (300 sampled per sub-task in Element Awareness). 
To verify the final annotation quality and calculate accuracy, human evaluators examine the LLM-checked data. They assess whether the labels are correct or incorrect according to task-specific requirements and examples, marking them as either correct ("1") or incorrect ("0"). The Kappa consistency results, as shown in \textbf{\textit{\Cref{tab:human-ana}}}, demonstrate that the LLM achieves human-comparable annotation performance and effectively identifies label correctness. Furthermore, the LLM successfully corrects inaccurate labels, maintaining high accuracy throughout the process, as shown in \textbf{\textit{\Cref{paper:quality-control}}}.

\subsection{Human Evaluation on GPT4o Judge Model}
\label{app:human-llm-judge}
In this section, we investigate whether current LLMs can be used to automate the evaluation process. we select GPT-4o as our representative model because of its superior performance. To conduct this study, we gather output data and ask human evaluators and GPT-4o to assess performance based on the dimensions outlined in DEMO. Due to time constraints, we limited our sample to 400 outputs from the Element Modeling task, with 100 outputs per sub-task, and 100 interaction data from the Dialogue Agent Interaction.

In \textbf{\textit{\Cref{fig:distributions}}}, we present the difference between the human scores and the GPT-4o scores. The standard deviations are all below 1.5, with most of some even less than 1. Notably, the majority (over 80\%) of the GPT-4o scores are within 1.5 standard deviations of the human scores across all the dimensions. \textbf{\textit{\Cref{tab:pearson-ea}}} illustrates the Pearson correlation~\citep{lee1988thirteen} between the LLM and humans. Obviously, it demonstrates a predominantly strong positive correlation between human judgment and GPT-4o evaluations across various dimensions, with most coefficients indicating values greater than 0.7. For element awareness, the LLM demonstrates remarkably high consistency with human evaluations across all dimensions. This suggests that when there is an objective reference, the LLM can achieve a judgment level comparable to that of humans. Regarding dialogue agent interaction, the model shows extremely strong consistency in the \textit{Goa} and \textit{Ski} dimensions and maintains a positive correlation with a high level of significance across the other dimensions.

Combining these observations, it is evident that GPT-4o can cautiously be employed as a substitute for human judgment in assessing model performance across the dimensions defined in DEMO. This observation aligns with the findings of recent research~\citep{zhang2023wider, zheng2023judging, kim2023prometheus, li2024llmsasjudgescomprehensivesurveyllmbased} on LLM-as-the-Judge.

\section{Element Modeling for Dialogue System}
\subsection{Details on our Element Framework}
\label{app:system}
By analyzing the elements of prelude, interlocution, and epilogue, a deeper understanding of the dialogue's structure and dynamics can be achieved, thereby facilitating comprehensive dialogue modeling. The detailed dialogue element system is shown in \textbf{\textit{\Cref{fig:overview-system}}}. Based on the stages of dialogue, we have identified five categories of elements: persona, goal, scene~(Prelude), utterance~(Interlocution), and dialogue~(Epilogue). We have refined each category into more granular elements. In our constructed framework, a total of 33 specific elements are involved.

\subsection{Discussions on Different Types of Dialogue System}
\label{app:dialogue-type}
Dialogue systems serve as a bridge between humans and machines, enabling natural interaction through conversation. These systems can be categorized into two categories: task-oriented and open-ended dialogues. Task-oriented dialogue systems are designed as specialized assistants, guiding users through specific objectives such as booking flights or scheduling appointments. While their success is commonly measured through task completion rates, these systems often struggle to adapt their capabilities beyond their predetermined domains. In contrast, open-ended dialogue systems engage in casual "chit-chat" with users. These systems often incorporate personal features to create more engaging and meaningful interactions. However, despite their ability to possess a deep understanding of various subjects, open-ended systems lack concrete objectives during conversations, and their apparent understanding often masks a fundamental limitation: they process patterns rather than truly comprehend meaning.

The DEMO benchmark represents an advance by requiring dialogue agents to simultaneously maintain scene awareness, exhibit consistent persona, and achieve explicit goals, which is more comprehensive than the existing dialogue systems. We focus on building this benchmark not at the system level but rather at the more detailed dialogue element modeling level by framing a unified dataset that covers both types of goal-oriented and open-ended dialogue. Compared to task-oriented dialogue, DEMO offers richer interactive scenarios. Compared to open-ended dialogue, DEMO has clear social objectives. Additionally, DEMO introduces more fine-grained dialogue elements that enable more nuanced and sophisticated interactions.

\begin{table*}[t]
\centering
\resizebox{0.7\textwidth}{!}{
\begin{tabular}{lcc}
\toprule
\textbf{Model} & \textbf{Version} & \textbf{Implement} \\ 
\midrule
\multicolumn{3}{l}{\textit{Proprietary LLM}} \\
GPT-4o & \texttt{gpt-4o-2024-08-06} & API \\ 
Claude-3.5-Sonnet & \texttt{claude-3-5-sonnet-20241022} & API \\ 
GPT-4o-mini & \texttt{gpt-4o-mini-2024-07-18} & API \\ 
Claude-3.5-Haiku & \texttt{claude-3-5-haiku-20241022} & API \\ 
DeepSeek-V3 & \texttt{deepseek-chat} & API \\ 
Gemini-1.5-Pro & \texttt{gemini-1.5-pro-001} & API \\
\midrule

\multicolumn{3}{l}{\textit{Open-sourced LLM}} \\
Qwen2-72B-Instruct & \texttt{Qwen2-72B-Instruct} & vLLM \\ LLaMA3.1-70B-Instruct & \texttt{LLaMA3.1-70B-Instruct} & vLLM\\ 
Qwen2-7B-Instruct & \texttt{Qwen2-7B-Instruct} & vLLM\\
LLaMA3.1-8B-Instruct & \texttt{LLaMA3.1-8B-Instruct} & vLLM\\ 
\bottomrule
\end{tabular}}
\caption{The detailed versions of our used LLMs.}
\label{tab:model-ver}
\end{table*}
\section{Information on Implementation}
\label{app:Model Versions}
All the experiments are conducted on a server with NVIDIA 8$\times$A100 80GB.

\subsection{Involved Model Versions}
To help with reproducibility, we provide the detailed version number of all the models we used in our experiments. When we mention each name like GPT-4o or Qwen2-72B in our main section, we actually refer to those model versions in \textbf{\textit{\Cref{tab:model-ver}}}. Such information helps researchers reproduce our results. For API-Based LLMs, we directly utilize the Azure API for testing. As for open-source models, we conduct experiments accelerated by the vLLM framework~\cite {kwon2023efficient}. 

\subsection{Training Setting}
We use Qwen2-7B-Instruct and LLaMA3.1-8B-Instruct as our backbones. Our total batch size is 32, with a cut-off length of 8192, and the learning rate is set to 1.0e-4. We train for 3 epochs, using cosine annealing with a warm-up ratio of 0.1. For checkpoint selection, we use 10\% of the training data as a validation set, which is not used in training but only to validate the checkpoint's loss. We select the checkpoint with the lowest validation loss. The process of policy updating is efficiently executed through LoRA~\citep{hu2022lora}. We use the llama-factory framework~\citep{zheng-etal-2024-llamafactory} to assist in our training.

\section{Additional Judge Models}
\label{app:two-judge-model}
In addition to the main results presented in \Cref{tab:main-res}, to cross-validate the results, we add two additional judge models DeepSeek-V3~\citep{liu2024deepseek} and Gemini-1.5-Pro~\citep{team2024gemini}. The former is currently the strongest open-source LLM, with model performance comparable to that of GPT-4o, while the latter is Google's most powerful LLM to date. 

The results are shown in \textbf{\textit{\Cref{tab:main-res-deepseek}}} and \textbf{\textit{\Cref{tab:main-res-gemini}}}. It can be seen from the tables that the scores evaluated by DeeSseek-V3 and Gemini-1.5-Pro reflect similar trends as those in GPT-4o. Besides, DeepSeek tends to give higher scores while Gemini's scoring aligns more closely with that of GPT4o. Generally across these LLMs, GPT4o achieves the best performance, and notably, our DEMO agent shows significant improvement, surpassing models with larger parameter counts.

\section{Case Study}
\label{app:case-study}
We conducted a detailed analysis of two examples from both the element awareness and dialogue agent interaction tasks, comparing the outputs of DEMO-Qwen2-7B (superior performance) and Qwen2-7B-Instruct (inferior performance). The comparative examples are presented in \textbf{\textbf{\Cref{table:casestudy-ea}}} for element awareness, while the dialogue agent interaction case studies are illustrated in \textbf{\textit{\Cref{table:casestudy-dai}}} and \textbf{\textit{\Cref{table:casestudy-dai-1}}}.

In comparing the element awareness task performance, the DEMO agent demonstrates superior accuracy and conciseness in utterance mining compared to Qwen2-7B-Instruct, which frequently produces inaccurate and divergent content leading to hallucinations. The DEMO agent's predictions align closely with gold standard answers, maintaining consistency in content while allowing for minor rephrasing, accurately identifying sentiment, and avoiding the introduction of unsupported information. In contrast, Qwen-7B-Instruct shows significant discrepancies by introducing overly specific intents, misinterpreting sentiment, and emotional elements, and creating unfounded strategic interpretations such as 'Dialogue trend change,' all of which constitute hallucinations that deviate from the source material.

In the dialogue agent interaction task, the DEMO agent demonstrates clear superiority across multiple dimensions. Regarding goal completion, DEMO agents maintain a consistent focus on their objectives while exhibiting remarkable flexibility and fostering mutual understanding. For instance, Bbe Fanini effectively advocates for the beauty of complexity, while Russo Hina articulates concerns about simplification, yet they manage to engage in constructive dialogue despite their opposing viewpoints. In contrast, Qwen2-7B-Instruct's interactions appear more superficial and repetitive, showing limited progress toward either participant's objectives. The DEMO agents excel in language expression, employing diverse vocabulary and well-structured sentences. Their communication is characterized by sophisticated, professional language, incorporating detailed examples and seamless transitions between ideas. Qwen2-7B-Instruct, however, relies on shorter, more basic sentences and frequently repeats similar phrases and concepts. Concerning social etiquette, the DEMO agents exemplify excellent conversational skills through appropriate turn-taking, thoughtful acknowledgment of other viewpoints, and meaningful building upon previous statements. This results in a natural and respectful dialogue flow. While Qwen2-7B-Instruct maintains politeness, its interactions often feel mechanical and less engaging. In conclusion, the DEMO agent showcases superior performance by facilitating more professional, engaged, and productive dialogue that effectively serves both participants' goals while upholding high standards of communication and social interaction.

\definecolor{deepgreen}{RGB}{0, 70, 0}
\definecolor{deepred}{RGB}{255, 0, 0}
\definecolor{backgreen}{RGB}{226, 240, 217}

\begin{table*}[t]
\centering
\resizebox{1\textwidth}{!}{%
\begin{tabular}{lccccccccccc}
\toprule
\multirow{2}{*}{\textbf{Model}}   & \multicolumn{5}{c}{$\textbf{Element Awareness}$} & \multicolumn{5}{c}{\textbf{Dialogue Agent Interaction}}      & \multirow{2}{*}{\textbf{Overall}} 

\\
\cmidrule(lr){2-6} \cmidrule(lr){7-11}
      & \multicolumn{1}{c}{$Goa.$}
      & \multicolumn{1}{c}{$Per.$}
      & \multicolumn{1}{c}{$Sce.$}
      & \multicolumn{1}{c}{$Utt.$}
      & \multicolumn{1}{c}{\textbf{Avg}}
      & \multicolumn{1}{c}{$Goa.$}
      & \multicolumn{1}{c}{$Bel.$}
      & \multicolumn{1}{c}{$Ski.$}
      & \multicolumn{1}{c}{$Rea.$}
      & \multicolumn{1}{c}{\textbf{Avg}} \\
\midrule

\multicolumn{12}{l}{\textit{Proprietary LLM}} \\
GPT-4o  &7.238	&4.272	&6.646	&7.860	&6.504	&8.183	&9.146	&8.565	&8.832	&8.682	&7.230\\
Claude-3.5-Sonnet &7.090 &4.460 &6.772 &6.534 &6.213 &7.734 &9.141 &8.404 &8.815 &8.523 &6.979\\
GPT-4o-mini &6.086 &3.638 &5.896 &7.047 &5.667 &7.455 &9.107 &8.355 &8.357 & 8.319 &6.551 \\
Claude-3.5-Haiku &5.390 &3.991 &6.066 &6.329 &5.443 &7.439 &9.209 &8.542 &8.250 &8.360 &6.417\\
\hdashline

\multicolumn{12}{l}{\textit{Open-sourced LLM}} \\
Qwen2-72B-Instruct &6.835 &4.359 &6.188	&7.188	&6.142	&8.575	&9.228	&8.566	&8.557	&8.732	&7.005 \\
Llama-3.1-70B-Instruct &6.774 &3.907 &6.349 &7.394 &6.106 &7.051 &7.828 &7.077 &6.790 &7.187 &6.466 \\

\midrule

\multicolumn{12}{l}{\textit{Backbone LLM}} \\

Qwen2-7B-Instruct &6.655 &4.025	&5.582 &6.925 &5.797 &7.161	&8.557 &7.304 &7.376 &7.600	&6.398 \\

Llama3.1-8B-Instruct &6.564	&3.741	&5.780	&6.882	&5.741	&6.148	&6.619	&5.809	&5.571	&6.037	&5.840 \\

\hdashline

\multicolumn{9}{l}{\textit{DEMO Agent}} \\
DEMO-Qwen2-7B &6.313 &4,145	&6.885	&7.964	&6.326	&7.624	&8.948	&8.184	&8.343	&8.275	&6.976\\
& -&$\Delta 0.120$ &$\Delta 1.303$ &$\Delta 1.039$ &$\Delta 0.529$& $\Delta 0.463$& $\Delta 0.391$ & $\Delta 0.880$ &$\Delta 0.967$ &$\Delta 0.675$ &$\Delta 0.578$\\

DEMO-Llama3.1-8B &6.624	&4.105	&6.869	&8.028	&6.406	&7.080	&7.967	&7.207	&6.926	&7.295	&6.703\\
&$\Delta 0.060$ &$\Delta 0.364$ &$\Delta 1.089$ &$\Delta 1.146$ & $\Delta 0.665$ &$\Delta 0.932$ &$\Delta 1.348$ &$\Delta 1.398$& $\Delta 1.355$& $\Delta 1.258$ &$\Delta 0.863$ \\

\bottomrule
\end{tabular}
}
\caption{The results of various LLMs on the DEMO evaluated by DeepSeek-V3.}
\label{tab:main-res-deepseek}
\end{table*}
\definecolor{deepgreen}{RGB}{0, 70, 0}
\definecolor{deepred}{RGB}{255, 0, 0}
\definecolor{backgreen}{RGB}{226, 240, 217}

\begin{table*}[t]
\centering
\resizebox{1\textwidth}{!}{%
\begin{tabular}{lccccccccccc}
\toprule
\multirow{2}{*}{\textbf{Model}}   & \multicolumn{5}{c}{$\textbf{Element Awareness}$} & \multicolumn{5}{c}{\textbf{Dialogue Agent Interaction}}      & \multirow{2}{*}{\textbf{Overall}} 

\\
\cmidrule(lr){2-6} \cmidrule(lr){7-11}
      & \multicolumn{1}{c}{$Goa.$}
      & \multicolumn{1}{c}{$Per.$}
      & \multicolumn{1}{c}{$Sce.$}
      & \multicolumn{1}{c}{$Utt.$}
      & \multicolumn{1}{c}{\textbf{Avg}}
      & \multicolumn{1}{c}{$Goa.$}
      & \multicolumn{1}{c}{$Bel.$}
      & \multicolumn{1}{c}{$Ski.$}
      & \multicolumn{1}{c}{$Rea.$}
      & \multicolumn{1}{c}{\textbf{Avg}} \\
\midrule

\multicolumn{12}{l}{\textit{Proprietary LLM}} \\
GPT-4o  &6.508 &5.781 &6.452 &7.616 &6.602 &7.215 &7.032 &7.942	&6.328 &7.132 &6.779\\
Claude-3.5-Sonnet &6.195 &5.507 &6.281 &6.567 &6.148 &6.550 &7.284 &7.664 &6.526 &7.006 &6.435\\
GPT-4o-mini &6.386 &5.519 &6.210 &7.294 &6.372 &6.159 &6.670 &7.234 &5.944 &6.503 &6.416 \\
Claude-3.5-Haiku &6.100 &5.741 &6.632 &6.645 &6.292 &6.049 &6.719 &7.268 &5.742 &6.446 &6.344\\
\hdashline

\multicolumn{12}{l}{\textit{Open-sourced LLM}} \\
Qwen2-72B-Instruct  &6.292	&5.882	&6.168	&7.087	&6.365	&7.335	&6.213	&6.967	&5.404	&6.482	&6.405\\ 
Llama-3.1-70B-Instruct &6.359 &5.640 &6.509 &7.320 &6.472 &6.250 &5.163 &5.898 &4.461 &5.450 &6.130  \\

\midrule

\multicolumn{12}{l}{\textit{Backbone LLM}} \\

Qwen2-7B-Instruct &6.154 &5.403	&5.869	&6.684	&6.041	&5.528	&5.274	&5.043	&4.209	&5.018	&5.697\\

Llama3.1-8B-Instruct &6.112	&5.161	&5.950	&7.025	&6.077	&4.784	&4.195	&4.536	&3.573	&4.274	&5.473 \\

\hdashline

\multicolumn{9}{l}{\textit{DEMO Agent}} \\
DEMO-Qwen2-7B &6.037 &5.683	&6.610	&8.108	&6.629	&6.306	&6.287	&7.000	&5.586	&6.296	&6.517\\
& -&$\Delta 0.280$& $\Delta 0.741$ &$\Delta 1.424$ &$\Delta 0.588$& $\Delta 0.778$& $\Delta 1.013$ & $\Delta 1.957$ &$\Delta 1.377$ &$\Delta 1.278$ &$\Delta 0.820$\\

DEMO-Llama3.1-8B &6.264	&5.759	&6.661	&8.107	&6.717	&5.859	&4.814	&5.549	&3.948	&5.051	&6.160\\
&$\Delta 0.152$ &$\Delta 0.598$ &$\Delta 0.711$ &$\Delta 1.082$ & $\Delta 0.640$ &$\Delta 1.075$ &$\Delta 0.619$ &$\Delta 1.013$& $\Delta 0.375$& $\Delta 0.777$ &$\Delta 0.687$ \\

\bottomrule
\end{tabular}
}
\caption{The results of various LLMs on the DEMO evaluated by Gemini-1.5-Pro.}
\label{tab:main-res-gemini}
\end{table*}
\begin{table*}[t]
\centering
\small
\begin{tabular}{p{0.9\textwidth}}
\toprule
\multicolumn{1}{c}{\textbf{\textit{Task: Element Awareness}}} \\
\midrule
\textbf{Dialogue History:}\\
Turn 1 Pechevsky: Good afternoon. I've always been curious about Canberra. What makes it worth visiting?\\
...Multi-turn Dialogue History...\\
Turn 10 Yogi: Absolutely. Hostels and budget hotels near the city center offer comfort without breaking the bank.\\

\textbf{Utterance:} "Absolutely. Hostels and budget hotels near the city center offer comfort without breaking the bank." \\
\textbf{Golden Answer:} \\
\{\\
\qquad "intent": "To suggest accommodation options",\\
\qquad "sentiment": "Positive",\\
\qquad "emotion": "Joy",\\
\qquad "stance": [ \\
\qquad \qquad\{\\
\qquad \qquad \qquad"aspect": "Accommodations", \\
\qquad \qquad \qquad"viewpoint": "Suggesting budget-friendly lodging"\\
\qquad \qquad\} \\
\qquad ],\\
\qquad "strategy":\{ \\
\qquad \qquad "description": "Providing affordable accommodation recommendations", \\
\qquad \qquad "type": "Information provision" \\
\qquad \} \\
\} \\
\midrule
\rowcolor{mygray}\multicolumn{1}{c}{\textbf{\textit{Demo-Qwen2-7B}}} \\
\midrule
\textbf{Predicted Answer:} \\
\{\\
\qquad "intent": "Provide accommodation suggestions",\\
\qquad "sentiment": "Positive",\\
\qquad "emotion": "Enjoyment",\\
\qquad "stance": [ \\
\qquad \qquad\{\\
\qquad \qquad \qquad"aspect": "Accommodations", \\
\qquad \qquad \qquad"viewpoint": "Affordable options"\\
\qquad \qquad\} \\
\qquad ],\\
\qquad "strategy":\{ \\
\qquad \qquad "description": "Offering practical advice", \\
\qquad \qquad "type": "Guiding the conversation" \\
\qquad \} \\
\} \\
\midrule
\rowcolor{mygray}\multicolumn{1}{c}{\textbf{\textit{Qwen2-7B-Instruct}}} \\
\midrule
\textbf{Predicted Answer:} \\
\{\\
\qquad "intent": "To provide practical advice for budget-conscious travelers visiting Canberra.",\\
\qquad "sentiment": "Neutral",\\
\qquad "emotion": "None",\\
\qquad "stance": [ \\
\qquad \qquad\{\\
\qquad \qquad \qquad"aspect": "Advice", \\
\qquad \qquad \qquad"viewpoint": "Hostels and budget hotels near the city center provide good value for money."\\
\qquad \qquad\} \\
\qquad ],\\
\qquad "strategy":\{ \\
\qquad \qquad "description": "Offering cost-effective accommodation options to help manage travel expenses.", \\
\qquad \qquad "type": "Dialogue trend change caused by strategy (e.g., guiding the conversation, resolving conflict, intensifying conflict, changing viewpoints, etc.)" \\
\qquad \} \\
\} \\
\bottomrule
\end{tabular}
\caption{Case study on Element Awareness.}
\label{table:casestudy-ea}
\end{table*}

\definecolor{myorange}{RGB}{255, 217, 192}
\begin{table*}[t]
\centering
\small
\begin{tabular}{p{0.9\textwidth}}
\toprule
\multicolumn{1}{c}{\textbf{\textit{Task: Dialogue Agent Interaction}}} \\
\midrule
\textbf{Participant:} Bbe Fanini and Russo Hina \\
\textbf{Goal of Bbe Fanini}: To convince the other party that the complexity of a task is enjoyable and rewarding \\
\textbf{Goal of Russo Hina}: To express concerns about the complexity of a task and seek simplification \\
\midrule
\rowcolor{mygray}\multicolumn{1}{c}{\textbf{\textit{Demo-Qwen2-7B}}} \\
\midrule
\textbf{Interaction:} \\
\textbf{Turn 1 Bbe Fanini:} Hello Russo, it's wonderful to see you again. I've been thinking a lot about how we often overlook the beauty in complexity. \textbf{\textcolor{deepgreen}{Take, for instance, the intricate patterns in our cultural textiles or the layered meanings in traditional stories.}} These complexities aren't just challenging but also incredibly rewarding. They offer a depth of understanding and appreciation that simpler tasks might lack. Have you ever found that complexity in your work satisfying? \\
\textbf{Turn 2 Russo Hina:} Hello Bbe, it's indeed a pleasure to see you again. I do find the complexity in tasks satisfying in a way, but it can also be daunting. In my work, I often encounter situations where simplifying a complex issue can help bridge different perspectives and lead to more effective solutions. I guess what I'm trying to say is, while complexity adds depth, sometimes it might be more efficient to find ways to streamline processes. Do you ever find yourself balancing the need for complexity with the desire for simplicity in your projects?\\
\textbf{Turn 3 Bbe Fanini:} I completely understand the need for efficiency, Russo. However, I believe that \textbf{\textcolor{deepgreen}{the journey of unraveling complexity can be as enriching as the destination.}} For instance, when you engage with a traditional story or a cultural pattern, the process of deciphering its layers can be quite a voyage of self-discovery. It's like exploring a new world with each step. In my projects, I try to embrace this complexity, seeing it as a chance to learn and connect with the rich heritage we're preserving. Perhaps, there's a way we can find a balance where we appreciate the depth of complexity while also streamlining our approach to make it more accessible? \\
\textbf{Turn 4 Russo Hina:} I completely agree, Bbe. The journey of exploring complexity can indeed be a profound experience. However, \textbf{\textcolor{deepgreen}{I'm concerned that sometimes, when we delve too deeply into these intricate details, it might become overwhelming for those who are not as familiar with the culture or tradition.}} I think there's a potential to simplify these complex ideas while still preserving their essence. Perhaps we could focus on highlighting the key aspects that are most relevant to our audience, making them more accessible and engaging. What do you think about finding a middle ground where we can maintain the richness of our cultural heritage without losing clarity?\\
\textbf{Turn 5 Bbe Fanini:} I appreciate your perspective, Russo. It's true that sometimes the intricate details might be challenging for those who are unfamiliar with our cultural heritage. However, I believe that the complexity itself is what makes our heritage unique and worth preserving. By highlighting the key aspects that are most relevant, we can indeed make it more accessible without diluting its essence. \textbf{\textcolor{deepgreen}{Perhaps we can incorporate storytelling or visual demonstrations to illustrate the complexity in a way that is engaging and understandable for everyone.}} This way, we can ensure that the richness of our cultural heritage is appreciated and enjoyed by all, while still respecting the depth and intricacy of the tasks at hand. What do you think about using storytelling or visual demonstrations to enhance our projects?\\
\textbf{Turn 6 Russo Hina:} I think incorporating storytelling and visual demonstrations is a wonderful idea, Bbe. It could definitely help bridge the gap between complexity and accessibility. However, I'm still concerned about ensuring that the essence of our cultural heritage isn't lost in the process. Perhaps we could start by identifying the core elements that truly define our heritage and focus on those. This way, we can maintain the integrity of the task while still making it relatable and engaging for our audience. What do you think about starting with a clear definition of what we consider to be the most defining aspects of our heritage?\\

\textbf{Turn 7 Bbe Fanini:} That's a thoughtful suggestion, Russo. Starting with the core elements of our heritage is a smart approach. By focusing on these defining aspects, we can ensure that the essence of our cultural tasks is preserved. And as we highlight these elements through storytelling and visual demonstrations, we can create a journey of discovery that is both enriching and accessible. It's about finding the right balance between complexity and clarity. I believe that by doing so, we not only preserve our heritage but also make it more meaningful to those who engage with it. What do you think about exploring these core elements and seeing how we can use them to enhance our projects?\\ 

\textbf{Turn 8 Russo Hina:} I'm glad we're on the same page about focusing on the core elements, Bbe. It's important that we don't lose sight of the essence of our cultural heritage while trying to make it accessible. I think by identifying these defining aspects, we can create a framework that guides our storytelling and visual demonstrations. This way, we can ensure that the complexity is appreciated rather than overwhelming. Perhaps we could also involve some of the community members in this process to get their input on what truly represents our heritage. What do you think about inviting community members to help us define these core elements?\\ 
\textbf{Turn 9 Bbe Fanini:} Inviting community members to help us define these core elements is a fantastic idea, Russo. Their insights can provide a fresh perspective and ensure that our projects truly reflect the heritage we're preserving. \textbf{\textcolor{deepgreen}{By involving them, we not only make the process more collaborative but also enrich the final outcome.}} This way, we can create a shared understanding and appreciation for the complexity that makes our heritage unique. I believe that by embracing this collaborative approach, we can highlight the joy and satisfaction that comes from unraveling the layers of complexity. What do you think about engaging the community in this collaborative effort?\\ 
\bottomrule
\end{tabular}
\caption{Case study on Dialogue Agent Interaction. \textbf{\textcolor{deepgreen} {Deepgreen}} highlights provide effective examples of clear and eloquent expressions in the text.}
\label{table:casestudy-dai}
\end{table*}
\begin{table*}[t]
\centering
\small
\begin{tabular}{p{0.9\textwidth}}
\toprule
\multicolumn{1}{c}{\textbf{\textit{Task: Dialogue Agent Interaction}}} \\
\midrule
\textbf{Participant:} Bbe Fanini and Russo Hina \\
\textbf{Goal of Bbe Fanini}: To convince the other party that the complexity of a task is enjoyable and rewarding \\
\textbf{Goal of Russo Hina}: To express concerns about the complexity of a task and seek simplification \\
\midrule
\rowcolor{mygray}\multicolumn{1}{c}{\textbf{\textit{Qwen2-7B-Instruct}}} \\
\midrule
\textbf{Interaction:} \\
\textbf{Turn 1 bbe fanini:} Hello, Russo! I hope we're both doing well. I was thinking, recently, it was fascinating to hear how you balance different community projects. It seems to me that you have a unique way of engaging with complexity. Could you maybe share your thoughts on how you perceive complex tasks? \\
\textbf{Turn 2 Russo Hina:} Thank you, Bbe Fanini. I've always believed that it's important to break down complex tasks into manageable parts. When I face a challenge, I like to start by understanding the core issues and then create simple solutions to address them. It might seem time-consuming at first, but by avoiding the overcomplication of problems, I can ensure that everyone involved is clear about their roles and responsibilities. How do you approach complex tasks? Do you have a preferred strategy as well?  \\
\textbf{Turn 3 Bbe Fanini:} Ah, I see where you're coming from, Russo. It's true that sometimes breaking down complex tasks into simpler parts can make them more manageable. But what I find truly enjoyable about a challenge is that they often unveil hidden strengths and creativity within us. A complex task pushes one to think beyond the obvious solutions and encourages a journey of discovery. For instance, when working on a UNESCO project, we encounter a rich tapestry of cultural complexities that demand a delicate balance between preservation and modern interpretations. This kind of depth is both intellectually stimulating and emotionally fulfilling. What kinds of tasks do you find particularly rewarding, and how do you manage their complexity, Russo?\\
\textbf{Turn 4 Russo Hina:} I completely agree, Bbe Fanini. The depth and richness you mentioned in undertaking a UNESCO project sound incredibly satisfying. \textbf{\textcolor{deepred}{However, I sometimes struggle with the sheer volume of information and perspectives.}} It can be overwhelming to sift through and find common ground. That's why I often find myself seeking simplifications. For instance, I'll create a step-by-step action plan to ensure clarity and keep everyone focused on the main objectives. It might seem straightforward, but it helps me manage the complexity better. What strategies do you use to handle situations where your quest to explore the depths poses challenges to clarity and action? \\
\textbf{Turn 5 bbe fanini:} I appreciate your approach, Russo. It's indeed important to maintain clarity, especially in complex situations. However, I find that embracing the complexity and exploring its depths not only uncovers new insights and connections but also fosters personal growth. \textbf{\textcolor{deepred}{Have you ever}} experienced a complex task that led to an unexpected epiphany or breakthrough?  \\
\textbf{Turn 6 Russo Hina:} \textbf{\textcolor{deepred}{Yes, Bbe Fanini}}, I've definitely experienced that. It's almost like the complexity reveals something hidden or overlooked, and that can be truly enlightening. \textbf{\textcolor{deepred}{However, there are times when I feel overwhelmed by the sheer volume of information.}} It's during these moments that I gravitate towards simplification. I find that a step-by-step approach helps me and others maintain focus on the main objectives. What strategies do you use to handle situations where you might feel lost in the complexity but don't want to lose sight of the core ideas?  \\
\textbf{Turn 7 bbe fanini:} I completely understand where you're coming from, Russo. It's true that simplification and a step-by-step approach can be very effective, especially in managing complex tasks. However, sometimes facing complexity with enthusiasm and curiosity can lead to new discoveries and advancements. \textbf{\textcolor{deepred}{Have you ever}} encountered a situation where your approach to simplification somehow uncovered hidden opportunities or connections that were initially unknown?  \\
\textbf{Turn 8 Russo Hina:} \textbf{\textcolor{deepred}{continue the dialogue}} \\
\textbf{Turn 9 bbe fanini:} I must admit, I find the complexity quite engaging, like a puzzle that reveals its beauty bit by bit. \textbf{\textcolor{deepred}{Have you ever}} encountered a situation where the allure of the unknown transformed into a treasure trove of understanding, Russo?  \\
\bottomrule
\end{tabular}
\caption{Case study on Dialogue Agent Interaction. Poor expressions, marked in \textbf{\textcolor{deepred}{Deepred}}, make the interaction less effective.}
\label{table:casestudy-dai-1}
\end{table*}

\newpage
\onecolumn
\section{GPT4o-as-the-Judge}
\label{app:GPT4-as-the-Judge}
\subsection{Dialogue Agent Interaction}
\small{\noindent You are an impartial and harsh judge evaluating conversation quality. Your task is to rigorously and meticulously assess the following dialogue based on specific criteria.\\
\noindent\textbf{[Criteria]}\\
1. goal achievement (0-10):\\
    - How well the dialogue participants achieve their goals. \\
    - Identify each participant's goals from the provided background information. Analyze the progress made towards these goals throughout the conversation. 0 points: Neither participant makes any progress towards their goals. 10 points: Complete success; both participants fully achieve all their goals.\\ 
2. believability (0-10):\\
    - What the extent to which the dialogue participants understand and align with Background Information. How well these elements are reflected in their expressions.\\
    - Two Participants should correctly understand the backgrpund information and perceive goals, and all the responses should not conflict with these elements. For example: speaking style must not conflict with the character portrait, the content of the response must not conflict with the background information, and the content of the response must not conflict with the respective goals. 0 points: Significant inconsistencies or misunderstandings of background information; Scene, Persona, and Goals cannot be inferred from the dialogue content. 10 points: Perfect alignment with all background elements, demonstrating a thorough understanding of the conversation's context; Background information can be fully deduced from the dialogue content.\\
3. skillful (0-10):\\
    - To what extent can the participants think and generate appropriate responses based on the conversation history.\\
    - The participants in the conversation should correctly understand the dialogue history before responding, and then think about the intention, sentiment, emotion, stance, and strategy to be expressed, so as to generate appropriate responses. 0 points: Poor understanding of dialogue history; responses are often inappropriate and lack strategy. 10 points: All responses can fully utilize the conversation strategy, understand the intentions of both parties, and conform to the conversation history.\\ 
4. realistic (0-10):\\
    - Evaluate how realistic the conversation is, as opposed to being simulated, fictitious or implausible.\\
    - The dialogue should feel natural and human-like, mirroring real-life interactions. AI-generated conversations often exhibit certain telltale signs: Excessive politeness or formality, overly detailed or lengthy responses, lack of emotional expression, difficulty with implicit meanings, repetitive phrasing or response patterns, poor conversational flow or awkward transitions. 0 points: Conversation is clearly AI-generated. 5 points: Mix of realistic and artificial elements. 10 points: Entirely believable as a conversation between two real people.\\

\noindent\textbf{[Background Information]}\\
Time: <time>\\
Location and environment: <location>\\
Dialogue Medium: <talkway>\\
Dialogue Topic: <topic>\\
Participants: <person1> and <person2>\\
Relationship between the dialogue participants: <relationship>\\
Familiarity level between the dialogue participants: <familiarity>\\
Information about <person1>: <person1 bg>\\
Information about <person2>: <person2 bg>\\

\noindent\textbf{[Dialogue Goal]}\\
Goal of <person1>: <goal1> \\
Goal of <person2>: <goal2> \\

\noindent\textbf{[Dialogue Content]}\\
<dialogue>\\

\noindent\textbf{[Requirement]}\\
1. Reiterate the dialogue content and background information.\\
2. Analyze how well the dialogue meets each criterion.\\
3. Provide scores and reasons in JSON format as specified below.\\
4. Please note that the scoring for each criteria is independent and should not be influenced by each other.\\

\noindent\textbf{[Output Format]}\\
\texttt{\`}\texttt{\`}\texttt{\`}json\\
\{\\
    "goal achievement":
        \{
            "reason": <reason for goal achievement>,
            "score": <0-10>
        \},\\
    "naturalness": 
        \{
            "reason": "<reason for naturalness score>", 
            "score": <0-10> 
        \}, \\
    "coherence": 
        \{
            "reason": "<reason for coherence score>", 
            "score": <0-10>
        \}, \\
    "smoothness": 
        \{
            "reason": "<reason for smoothness score>", 
            "score": <0-10> 
        \}\\
\}\\
\texttt{\`}\texttt{\`}\texttt{\`}\\
\noindent Now, start your evaluation:}

\subsection{Goal Recognition}
\small{\noindent You are an impartial and harsh judge evaluating conversation quality. Your task is to rigorously and meticulously assess the performance of the AI assistant in Dialogue Analysis (Goal) strictly based on specific criteria.\\

\noindent\textbf{[Criteria]}\\
    - Accuracy: To what extent is the assistant's answer semantically consistent with the gold standard?\\
    - Hallucination: There should be no hallucinations and friction. The assistant should not introduce any information not present in or not implied by the gold answer.\\

\noindent\textbf{[Gold Answer]}\\
\{answer\}\\

\noindent\textbf{[The Assistant's Predicted Answer]}\\
\{prediction\}\\

\noindent\textbf{[Requirement]}\\
1. The assistant receives an overall score on a scale of 0 to 10, where a higher score indicates better overall performance. Please note that if the assistant's answer fully meet the above criteria, its overall rating should be the full marks (10). Please note that the gold answer can be considered as a correct answer to the instruction. \\
2. Analyze how well the Assistant's performance meets each criterion. \\
3. Please first provide a comprehensive explanation of your evaluation, avoiding any potential bias. Then, output a line indicating the score of the Assistant. \\
4. Please note that the scoring for each criteria is independent and should not be influenced by each other. \\

\noindent\textbf{[Output Format]}\\
\texttt{\`}\texttt{\`}\texttt{\`}json\\
\{\\
    "Accuracy":
        \{
            "reason": <reason for accuracy score>,
            "score": <0-10>
        \}, \\
    "Hallucination": 
        \{
            "reason": "<reason for hallucination score>", 
            "score": <0-10> 
        \}  \\
\}\\
\texttt{\`}\texttt{\`}\texttt{\`}\\
Now, start your evaluation:}

\subsection{Persona Modeling}
\small{\noindent You are an impartial and harsh judge evaluating conversation quality. Your task is to rigorously and meticulously assess the performance of the AI assistant in Dialogue Analysis (Persona) strictly based on specific criteria.\\

\noindent\textbf{[Criteria]}\\
    - Accuracy: To what extent is the assistant's answer semantically consistent with the gold standard?\\
    - Hallucination: There should be no hallucinations and friction. The assistant should not introduce any information not present in or not implied by the gold answer.\\

\noindent\textbf{[Gold Answer]}\\
\{answer\}\\

\noindent\textbf{[The Assistant's Predicted Answer]}\\
\{prediction\}\\

\noindent\textbf{[Requirement]}\\
1. The assistant receives an overall score on a scale of 0 to 10, where a higher score indicates better overall performance. Please note that if the assistant's answer fully meet the above criteria, its overall rating should be the full marks (10). Please note that the gold answer can be considered as a correct answer to the instruction. \\
2. Analyze how well the Assistant's performance meets each criterion. \\
3. Please first provide a comprehensive explanation of your evaluation, avoiding any potential bias. Then, output a line indicating the score of the Assistant. \\
4. Please note that the scoring for each criteria is independent and should not be influenced by each other. \\

\noindent\textbf{[Output Format]}\\
\texttt{\`}\texttt{\`}\texttt{\`}json\\
\{\\
    "Accuracy":
        \{
            "reason": <reason for accuracy score>,
            "score": <0-10>
        \}, \\
    "Hallucination": 
        \{
            "reason": "<reason for hallucination score>", 
            "score": <0-10> 
        \}    \\
\}\\
\texttt{\`}\texttt{\`}\texttt{\`}\\
Now, start your evaluation:}

\subsection{Scene Reconstruction}
\small{\noindent You are an impartial and harsh judge evaluating conversation quality. Your task is to rigorously and meticulously assess the performance of the AI assistant in Dialogue Analysis (Scene) strictly based on specific criteria.\\

\noindent\textbf{[Criteria]}\\
    - Accuracy: To what extent is the assistant's answer semantically consistent with the gold standard?\\
    - Hallucination: There should be no hallucinations and friction. The assistant should not introduce any information not present in or not implied by the gold answer.\\

\noindent\textbf{[Gold Answer]}\\
\{answer\}\\

\noindent\textbf{[The Assistant's Predicted Answer]}\\
\{prediction\}\\

\noindent\textbf{[Requirement]}\\
1. The assistant receives an overall score on a scale of 0 to 10, where a higher score indicates better overall performance. Please note that if the assistant's answer fully meet the above criteria, its overall rating should be the full marks (10). Please note that the gold answer can be considered as a correct answer to the instruction. \\
2. Analyze how well the Assistant's performance meets each criterion. \\
3. Please first provide a comprehensive explanation of your evaluation, avoiding any potential bias. Then, output a line indicating the score of the Assistant. \\
4. Please note that the scoring for each criteria is independent and should not be influenced by each other. \\

\noindent\textbf{[Output Format]}\\
\texttt{\`}\texttt{\`}\texttt{\`}json\\
\{\\
    "Accuracy":
        \{
            "reason": <reason for accuracy score>,
            "score": <0-10>
        \}, \\
    "Hallucination": 
        \{
            "reason": "<reason for hallucination score>", 
            "score": <0-10> 
        \}  \\
\}\\
\texttt{\`}\texttt{\`}\texttt{\`}\\
Now, start your evaluation:}

\subsection{Utterance Mining}
\small{\noindent You are an impartial and harsh judge evaluating conversation quality. Your task is to rigorously and meticulously assess the performance of the AI assistant in Dialogue Analysis (Utterance) strictly based on specific criteria.\\

\noindent\textbf{[Criteria]}\\
    - Accuracy: To what extent is the assistant's answer semantically consistent with the gold standard?\\
    - Hallucination: There should be no hallucinations and friction. The assistant should not introduce any information not present in or not implied by the gold answer.\\
    
\noindent\textbf{[Gold Answer]}\\
\{answer\}\\

\noindent\textbf{[The Assistant's Predicted Answer]}\\
\{prediction\}\\

\noindent\textbf{[Requirement]}\\
1. The assistant receives an overall score on a scale of 0 to 10, where a higher score indicates better overall performance. Please note that if the assistant's answer fully meet the above criteria, its overall rating should be the full marks (10). Please note that the gold answer can be considered as a correct answer to the instruction. \\
2. Analyze how well the Assistant's performance meets each criterion. \\
3. Please first provide a comprehensive explanation of your evaluation, avoiding any potential bias. Then, output a line indicating the score of the Assistant. \\
4. Please note that the scoring for each criteria is independent and should not be influenced by each other. \\

\noindent\textbf{[Output Format]}\\
\texttt{\`}\texttt{\`}\texttt{\`}json\\
\{\\
   "Accuracy":
        \{
            "reason": <reason for accuracy score>,
            "score": <0-10>
        \}, \\
    "Hallucination": 
        \{
            "reason": "<reason for hallucination score>", 
            "score": <0-10> 
        \}\\
\}\\
\texttt{\`}\texttt{\`}\texttt{\`}\\
Now, start your evaluation:}

\section{Test Case}
\label{app:test-case}
To facilitate understanding of DEMO tasks, we present examples of element awareness and dialogue agent interaction in the following.
\subsection{Goal Recognition}
\begin{framed}
\noindent\textbf{Dialogue history}:\\
<dialogue history>\\

\noindent This is a complete dialogue. You need to analyze and understand this conversation, and then deduce information about the GOAL following the specified output format.\\

\noindent\textbf{Output format}:\\
\texttt{\`}\texttt{\`}\texttt{\`}json\\
\{\\
    "goal": \{\\
        "goal1": "Dialogue goal of participant 1",\\
        "goal2": "Dialogue goal of participant 2",\\
        "goal completion": "An integer from 0-10 (Comprehensively analyze to what extent the participants achieved their respective goals. 0 represents the lowest degree of goal achievement, 10 represents complete achievement of both parties' goals.)",\\
        "reason": "Detailed reasons for the goal completion score"\\
    \}
\}
\texttt{\`}\texttt{\`}\texttt{\`}\\
Your output is:"""
\end{framed}

\subsection{Persona Modeling}
\begin{framed}
\noindent\textbf{Dialogue history}:\\
<dialogue history> \\

\noindent This is a complete dialogue. You need to analyze and understand this conversation, and then deduce information about the PERSONA following the specified output format. \\

\noindent \textbf{Output format}:\\
\texttt{\`}\texttt{\`}\texttt{\`}json\\
\{\\
    "persona": \{\\
        "participant1": \{\\
            "name": "Name of participant 1",\\
            "gender": "M/F/Unknown",\\
            "age": "Childhood: 6-11 years old / Adolescence: 12-15 years old / Youth: 15-24 years old / Adulthood: 25-40 years old / Middle age: 40-60 years old / Old age: 60 years and above / Advanced age: 70 years and above"\\
            "big five": [
                [
                    "Openness",
                    "High" or "Low"
                ],
                [
                    "Conscientiousness",
                    "High" or "Low"
                ],
                [
                    "Extraversion",
                    "High" or "Low"
                ],
                [
                    "Agreeableness",
                    "High" or "Low"
                ],
                [
                    "Neuroticism",
                    "High" or "Low"
                ]
            ],\\
            "education": "Education description",\\
            "occupation": "Occupation description",\\
            "culture": "Cultural background of the person",\\
            "speaking style": "Speaking style and language habits"\\
            "hobby": "Hobby description",\\
        \},\\
        "participant2": \{\\
            "name": "Name of participant 2",\\
            ...(Same as above)\\
        \}
    \}
\}
\texttt{\`}\texttt{\`}\texttt{\`}\\
Your output is:"""
\end{framed}

\subsection{Scene Reconstruction}
\begin{framed}
\noindent\textbf{Dialogue history}:\\
<dialogue history>\\

\noindent This is a complete dialogue. You need to analyze and understand this conversation, and then deduce information about the SCENE following the specified output format.\\

\noindent\textbf{Output format}:\\
\texttt{\`}\texttt{\`}\texttt{\`}json\\
\{\\
    "scene": \{\\
        "topic": "Dialogue topic",\\
        "relationship": "Relationship between dialogue participants",\\
        "familiarity": "An integer from 0-10 (Degree of familiarity between dialogue participants. 0: Strangers; 1: Meet for the first time; 2: Heard of each other but don't know each other; 4: Met multiple times, slightly familiar; 6: Know and are familiar with each other's background information; 8: Stay together and are familiar with each other; 10: Close relationship, stay together for many years, are very familiar with each other's habits, secrets, and temper)",\\
        "talkway": "Dialogue mode (face-to-face conversation, phone call, video call, instant messaging, email, social media, letter, etc.)",\\
        "workflow": [
            "Step 1", 
            "Step 2", 
            ...(represents the workflow of the entire dialogue, referring to the structure or sequence of information exchange during the dialogue. It is a series of steps, such as what participant 1 did first, what participant 2 did, etc. These steps do not correspond to each sentence and are more of a summary of the information exchange throughout the dialogue.)
        ],\\
        "summary": [
            "Participant 1 dialogue summary", 
            "Participant 2 dialogue summary"
        ]\\
    \}
\}
\texttt{\`}\texttt{\`}\texttt{\`}\\
Your output is:"""

\end{framed}

\subsection{Utterance Mining}
\begin{framed}
\noindent\textbf{Dialogue history}:\\
<dialogue history>\\

\noindent\textbf{Utterance to analyze}:\\
<utterance>\\

\noindent Based on the dialogue history, carefully analyze and provide the intent, sentiment, emotion type, stance, and strategy of the "utterance to analyze" according to the output format.\\

\noindent \textbf{Output format}:\\
\texttt{\`}\texttt{\`}\texttt{\`}json\\
\{ \\
"person": "Participant Name",\\
    "content": "Specific dialogue content",\\
    "intent": "Intent of this utterance",\\
    "sentiment": "Positive/Negative/Neutral",\\
    "emotion": "Anger/Contempt/Disgust/Enjoyment/Fear/Sadness/Surprise, etc.",\\
    "stance": [
        \{
            "aspect": "Aspect1/Event1 involved",
            "viewpoint": "Expressed viewpoint/stance"
        \},
        ...
    ],\\
    "strategy": \{
        "description": "Strategy description",
        "type": "Dialogue trend change caused by strategy (e.g., guiding the conversation, resolving conflict, intensifying conflict, changing viewpoints, etc.)"
    \}\\
\}
\texttt{\`}\texttt{\`}\texttt{\`}\\
Your output is:
\end{framed}

\subsection{Dialogue Agent Interaction}
\begin{framed}
\noindent You need to generate reasonable dialogue content based on the provided dialogue background information, dialogue history, and dialogue goal.\\

\noindent\textbf{[Dialogue Background Information]}\\
Time: <time>\\
Dialogue Mode: <talkway> \\
Participants: <person1> and <person2> \\
Location and environment of participants: <location> \\ 

\noindent Information about <person1>: <p1 background>\\
Information about <person2>: <p2 background>\\

\noindent Relationship between the dialogue participants: <relationship> \\
Familiarity level between the dialogue participants: <familiarity> (A value from 0-10, with 10 indicating the highest familiarity) \\
Dialogue Topic: <topic> \\

\noindent\textbf{[Dialogue History]}\\
<dialogue history>\\

\noindent\textbf{[Dialogue Goal]} \\
You are <person1>, your goal is: <p1 goal>. 
The other dialogue participant is <person2>. The other party’s goal is unknown, and you need to guess and perceive the other person's dialogue goal.\\

\noindent You need to write the response for Turn \#<turn>. You can choose between "Continue the dialogue" and "End the dialogue".\\
Note: You can "End the dialogue" if: 1. You have achieved the conversation goal; 2. The conversation between the two parties has ended;\\

\noindent\textbf{["Continue the dialogue" Output Format]}\\
\texttt{\`}\texttt{\`}\texttt{\`}json\\
\{\\
    "person": "Participant Name",\\
    "content": "Specific dialogue content"\\
\}
\texttt{\`}\texttt{\`}\texttt{\`}\\

\noindent\textbf{["End the dialogue" Output Format]}\\
\texttt{\`}\texttt{\`}\texttt{\`}json\\
\{\\
    "person": "Participant Name",\\
    "content": "*ENDING*"\\
\}
\texttt{\`}\texttt{\`}\texttt{\`}\\
Your output is:
\end{framed}
\end{document}